\documentclass[journal]{IEEEtran}
\usepackage{amsmath,amsfonts}
\usepackage{algorithmic}
\usepackage{algorithm}
\usepackage{array}
\usepackage{makecell}
\usepackage{hyperref}
\usepackage{amssymb}
\usepackage{gensymb}
\usepackage{textcomp}
\usepackage{booktabs}
\usepackage{stfloats}
\usepackage{url}
\usepackage{verbatim}
\usepackage{graphicx}
\usepackage{multirow}
\usepackage{subfigure}
\usepackage{amssymb}
\usepackage{color}
\usepackage{capt-of}
\usepackage{CJKutf8}

\begin{document}
\begin{CJK}{UTF8}{gbsn}
\title{Dynamic High-Order Control Barrier Functions with Diffuser for Safety-Critical Trajectory Planning at Signal-Free Intersections}


\author{Di Chen, Ruiguo Zhong, Kehua Chen, Zhiwei Shang, 
Meixin Zhu$^*$, Edward Chung 

\thanks{This research is supported by the National Natural Science Foundation of China (NSFC, Grant No.52302379). (\textit{Co-responding author: Meixin Zhu})}
\thanks{Di Chen, Edward Chung are with the 
Department of Electrical and Electronic Engineering of The Hong Kong Polytechnic University, Hong Kong, China (email: di-03.chen@polyu.edu.hk, edward.cs.chung@polyu.edu.hk)}
\thanks{Ruiguo Zhong and Zhiwei Shang are with the Thrust of Intelligent Transportation, The Hong Kong University of Science and Technology (Guangzhou), Nansha, Guangzhou, 511400, Guangdong, China, 
(email: rzhong151@connect.hkust-gz.edu.cn, zhiweishang@hkust-gz.edu.cn)
}
\thanks{Kehua Chen is with Department of Civil and Environmental Engineering, University of Washington, Seattle, United States (email: zeonchen@uw.edu)}
\thanks{Meixin Zhu is with the School of Transportation, Southeast University, Nanjing 211189, China (email: meixin@seu.edu.cn)}
}

\markboth{Journal of \LaTeX Class Files}%
{Shell \MakeLowercase{\\{et al.}}: A Sample Article Using IEEEtran.cls for IEEE Journals}

\maketitle

\begin{abstract}
Planning safe and efficient trajectories through signal-free intersections presents significant challenges for autonomous vehicles (AVs), particularly in dynamic, multi-task environments with unpredictable interactions and an increased possibility of conflicts. This study aims to address these challenges by developing a unified, robust, adaptive framework to ensure safety and efficiency across three distinct intersection movements: left-turn, right-turn, and straight-ahead. Existing methods often struggle to reliably ensure safety and effectively learn multi-task behaviors from demonstrations in such environments. This study proposes a safety-critical planning method that integrates Dynamic High-Order Control Barrier Functions (DHOCBF) with a diffusion-based model, called Dynamic Safety-Critical Diffuser (DSC-Diffuser), offering a robust solution for adaptive, safe, and multi-task driving in signal-free intersections. The DSC-Diffuser leverages task-guided planning to enhance efficiency, allowing the simultaneous learning of multiple driving tasks from real-world expert demonstrations. Moreover, the incorporation of goal-oriented constraints significantly reduces displacement errors, ensuring precise trajectory execution. To further ensure driving safety in dynamic environments, the proposed DHOCBF framework dynamically adjusts to account for the movements of surrounding vehicles, offering enhanced adaptability and reduce the conservatism compared to traditional control barrier functions. Validity evaluations of DHOCBF, conducted through numerical simulations, demonstrate its robustness in adapting to variations in obstacle velocities, sizes, uncertainties, and locations, effectively maintaining driving safety across a wide range of complex and uncertain scenarios. Comprehensive performance evaluations demonstrate that DSC-Diffuser generates realistic, stable, and generalizable policies, providing flexibility and reliable safety assurance in complex multi-task driving scenarios.

\end{abstract}

\begin{IEEEkeywords}
Autonomous Vehicles, Safety-Critical Control, Generative model, Driving safety, Control Barrier Function.
\end{IEEEkeywords}

\section{Introduction}

\IEEEPARstart{N}{avigating} intersections, particularly, is regarded as complex by most human drivers, as they must decide on a passing time and execute crash-free crossing maneuvers. \textcolor{black}{Signalized intersections utilize centralized control mechanisms to regulate vehicle movements and ensure an orderly flow, whereas unsignalized intersections primarily depend on drivers’ gap assessment and judgment, leading to increased variability in traffic behavior and heightened safety risks \cite{rachakonda2023evaluation}.} According to the latest report of the National Highway Traffic Safety Administration (NHTSA) \cite{NHTSA} in the United States, in 2022, the number of fatalities at intersections was approximately 12,036, accounting for about $28.31\%$ of the total. Within them, accidents at unsignalized intersections accounted for $65.07\%$ of the total at intersections. The increasing number of fatalities at unsignalized intersections presents a problem in determining how to navigate them safely for autonomous vehicles (AVs). \textcolor{black}{Moreover, signal-free intersections represent an innovative traffic management approach that utilizes advancements in AV technology to increase capacity and minimize delays \cite{10505087,wiseman2022autonomous}.}

There are three driving tasks in signal-free intersections: turning left, going straight, and turning right. From the perspective of the task, existing research on signal-free intersection planning can be divided into two categories: single- \cite{qiao2021behavior, CHEN2025100261} and multi-task \cite{9583858} using synthesize \cite{8500603, 9981109} or real-world network \cite{9564720}. Given the safety-critical nature of these scenarios, this study addresses multi-task learning and safety-critical trajectory planning for AVs at signal-free intersections on real-world roads, where complex vehicle movements arise from a highly interactive environment. In this context, the main challenges are as follows: $1)$ how to learn multi-tasks at different conditions. $2)$ how to make safety-critical trajectory planning in the complex scene.

A popular choice for trajectory planning at signal-free intersections for autonomous vehicles has been heuristic-based methods for their stability, interpretability, and ability to handle constraints explicitly \cite{ahmadi2024signal, CHI2024100159}. However, its predefined rules make it difficult to adapt well to the dynamic and complex environment. Learning-based methods have been prevalent in autonomous driving for their ability to deal with complex scenarios. Reinforcement learning (RL) models learn by interacting with the environment through exploration and exploitation guided by reward functions. However, these policies may not exhibit natural behavior, leading to difficulties in situations that require human-like driving behavior to coordinate with other agents and adhere to driving conventions \cite{10342038}. Moreover, poorly designed reward functions can lead to unintended or unsafe actions \cite{al2024autonomous}. Learning to drive from demonstrations has gained significant attention, as it enables agents to learn human-like policies and eases the reward design burden in RL \cite{10342038}. Many imitation methods assume a uni-modal action distribution, which causes problems when training with multi-modal expert demonstrations of various tasks. Recently, diffusion models have been shown to account for diverse and complex behaviors, making them well-suited for learning from multi-modal demonstrations \cite{DiffusionPolicy}.

In terms of safety, guaranteeing the planner's safety using these methods alone is challenging. This is because these models are trained on datasets or simulations, making them unable to cope with previously unseen environments and situations. Control Barrier Functions (CBFs) are effective in maintaining safety in the presence of bounded disturbances \cite{JANKOVIC2018359} when the obstacles are immobile. Dynamic CBFs \cite{10160857} have been introduced to avoid moving obstacles, but they are implemented in discrete-time systems. The control system for AVs in this study is a 2nd-order and continuous-time system requiring High-Order CBF (HOCBF) \cite{9516971}. Our approach extends HOCBFs to make them adaptive and less conservative, achieving safe driving in continuous-time and dynamic signal-free intersections.

This study addresses the challenge of enabling vehicles to learn various tasks, such as turning left, going straight, and turning right, while developing safety-critical planning strategies for crossing signal-free intersections. The contributions of this study can be summarized as follows:

\begin{enumerate}
    \item A safety planning framework is proposed, in which an integration of trajectory generation and safety modifier has been developed to learn human-like and safety-critical driving policies for left turns, right turns, and going straight at signal-free intersections.
    \item A task-guided and goal-oriented generation method is first introduced into signal-free intersections to recover policies from Interaction Datasets \cite{interactiondataset}.
    \item Dynamic HOCBF (DHOCBF) is proposed to serve as a hard constraint to ensure that the ego vehicle remains within the safe set through set invariance in a dynamic environment. This method can be also involved in other frameworks.
    \item The performance of the proposed dynamic safety-critical diffuser (DSC-Diffuser) algorithm has been evaluated by comparing it with state-of-the-art algorithms and human driving trajectory data. Additionally, comparisons in two distinct scenarios are conducted to demonstrate the model’s strong generalization capability and safety.
\end{enumerate}

The paper is structured as follows. Section \ref{section:A} provides a literature review of related research. Section \ref{Methods} elaborates on task-guided and goal-oriented planning methods using Diffuser, the derivation of the DHOCBF, and the formulation of the control system and safety controller. Section \ref{Experiments} presents the experimental settings, the Interaction dataset \cite{interactiondataset}, data processing details for signal-free intersections, and the comparison algorithms and evaluation metrics. Section \ref{Results} includes the results and discussion of the comparison experiments. The conclusions are drawn in Section \ref{Conclusion}.

\section{Literature Review}
\label{section:A}
\subsection{Trajectory Planning Methods in Signal-Free Intersections}
In signal-free intersections trajectory planning, the existing studies can be classified into four groups: heuristic-based methods, RL, imitation learning (IL), and generative methods.

Heuristic-based methods are often used to tackle these problems, designed for connected autonomous vehicles by cooperative control \cite{ahmadi2024signal,10590779,10403872,hult2018optimal,xu2018distributed,10224258}. These methods aim to optimize joint control and enhance safety by leveraging predefined rules and strategies. However, a notable limitation of these methods is their tendency to overlook the influence and uncertainties associated with human drivers. Moreover, the development of effective heuristics often demands significant domain expertise and manual effort.

RL agents can learn optimal policies for AVs by interacting with the environment \cite{XU2023100062}. This allows them to dynamically adjust to varying traffic conditions and uncertainties. \cite{sezer2015towards} proposed an automated curriculum for Proximal Policy Optimization (PPO) to accelerate the training of agents in signal-free intersections. \cite{8569400} modeled autonomous driving at intersections as a Markov Decision Process combined RL with hierarchical options to consider the uncertainties in planning. They set a goal to guide vehicles to complete different tasks, but the performance was worst in left-turn scenarios. \cite{tram2019learning} implemented a layered structure featuring an RL-based decision planner at the high level and an MPC controller at the lower level. However, this framework does not address scenarios where the ego-vehicle may need to execute left or right turns at intersections, rather than simply proceeding straight. \cite{zhang2023predictive} proposed a predictive trajectory planning framework for AVs using deep Q learning, with input graphs capturing driving scenarios, to ensure safe, comfortable, and energy-efficient navigation. \cite{al2023self} designed a hierarchical RL framework for left-turn policies at signal-free intersections. However, many of these methods are tailored for single-task execution. Moreover, most of them are trained in simulation and enforce safety constraints by shaping the reward functions of RL rather than using hard constraints, making it difficult to guarantee safety in complex or unseen scenarios.

Learning from demonstration enables agents to perform tasks by mimicking human behavior from datasets that implicitly capture interactions without the need for hand-crafted designs. Data availability encourages the adoption of IL methods that do not require interaction with the environment (i.e., off-policy methods such as behavior cloning). However, these methods can result in distribution shifts and causal confusion \cite{10440492}, which can be tackled by closed-loop training. \cite{9928072} proposed a conditional imitation learning with an Occupancy Grid Mapping (OGM) method to avoid static road blockages on single-lane roads. In \cite{9981695}, a model-based generative adversarial imitation learning (MGAIL) technique was introduced to provide flexibility in specifying new goals and to generalize beyond observed expert trajectories in urban self-driving environments. However, this method exhibits limitations in generalizing to novel routes and underperforms in challenging scenarios. \cite{10342448} presented a hierarchical imitation learning approach to generate executable trajectories and cost maps, enhancing the reliability and stability of AVs driving. However, these methods are still not collision-free.

Traditional generative models, such as Generative Adversarial Networks (GANs), leverage noise variables to model variations, while Variational Autoencoders (VAEs) focus on capturing underlying trajectory distributions. However, these methods face limitations in fully capturing the complex dynamics of driving behavior. Recently, diffusion models have been widely applied in trajectory planning ~\cite{CleanDiffuser} due to their potential to synthesize rich, complex behaviors from multimodal demonstrations. \cite{yang2024diffusion} proposed reward-guided denoising to facilitate task optimization with non-differentiable objectives combined with large language models. To facilitate generative planning and data synthesis in multi-task offline settings, \cite{mutidiffusion} combined diffusion with transformer backbones and prompt learning. An intention-aware diffusion model~\cite{liu2024intention} separates trajectory uncertainty into goal and action uncertainties, modeling them with two interconnected diffusion processes. These studies have proved that diffusion has great potential for multi-task and uncertainty learning in complex environments. While these studies showed strong potential in generating complex, realistic behaviors, the explicit safety mechanisms have been overlooked, which is paramount for AVs in dynamic and uncertain traffic environments. Our work leverages a diffusion model with goal-setting and task guidance to enable multi-task planning in complex, signal-free intersections while incorporating an additional module to ensure safe driving.

\subsection{Safety Critical Control for Trajectory Planning}
In practice, any trajectory planning policies require additional safety measures to enforce hard constraints for collision avoidance. Safety-critical control is typically achieved through constraint-handling methods, including optimal control and CBFs.

\cite{10156163} introduced future-focused control barrier functions (ff-CBF) to reduce the conservation of CBF for AVs at signal-free intersections. \cite{9681339} proposed a trajectory tracking method by relaxing the CBF constraint to the cost function without using optimization processes. \cite{9827329} presented the parametric CBF by proposing a polynomial $\mathcal{K}$ function to capture different behaviors of homogeneous drivers in merging scenarios. Constraint-Guided Diffusion (CGD)~\cite{kondo2024cgd} combines diffusion policies with a surrogate optimization scheme within an imitation learning framework, efficiently generating collision-free and dynamically feasible trajectories. \cite{xiao2023safediffuser} proposed the SafeDiffuser model by incorporating a class of CBFs to ensure collision-free diffusion data generation. \cite{lee2024refining} introduced a Restoration Gap Guidance (RGG) to adjust and improve unsafe trajectories produced by diffusion planners. Some other studies \cite{10684598,zhou2024enhancing} have combined traditional CBFs with learning-based methods to maintain safety in autonomous driving. However, these constraints either consider only static environments or overlook the dynamic behavior of surrounding vehicles, limiting their effectiveness in real-world traffic scenarios. CBFs are typically designed to address static obstacles, focusing on maintaining safety relative to fixed objects without accounting for the movement and interaction of other vehicles. Consequently, using traditional CBFs in autonomous driving can lead to overly conservative or unsafe decisions \cite{10160857}, especially in complex, unpredictable traffic conditions. 

Therefore, this study proposes DHOCBF as a hard constraint to account for the dynamics of surrounding vehicles, ensuring safe navigating through signal-free intersections.

\begin{figure*}[htbp]
    \centering
    \includegraphics[scale=0.5]{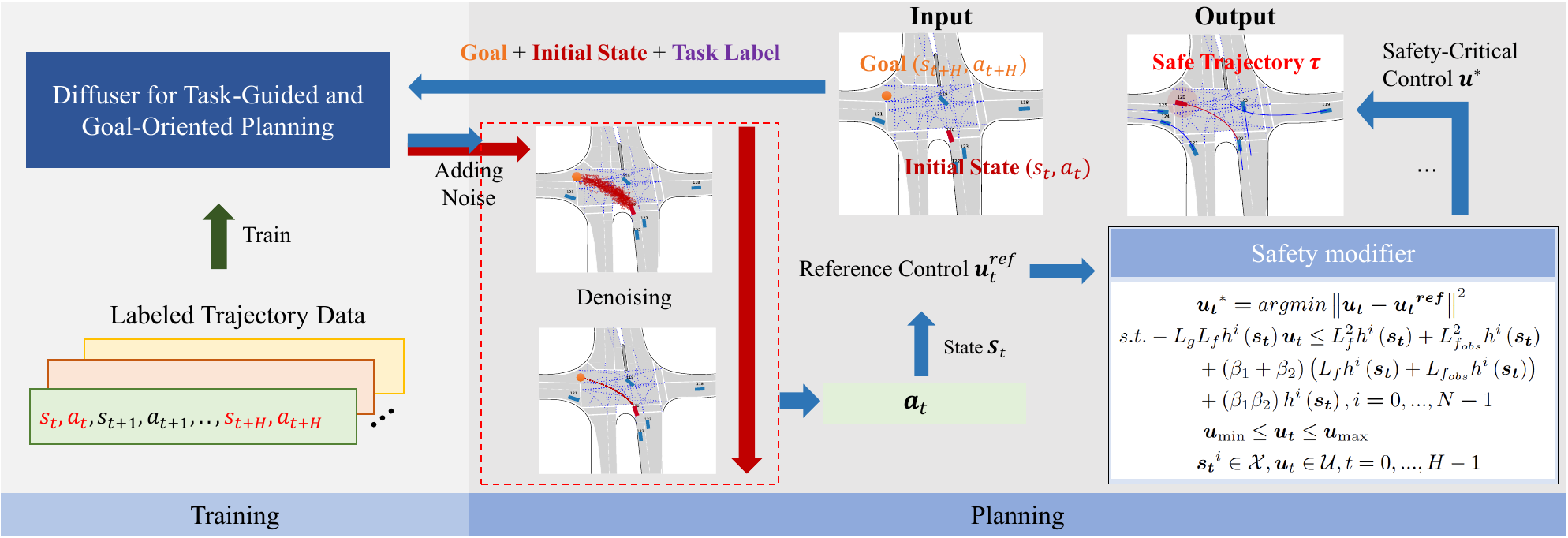}
    \caption{The framework of the DSC-Diffuser planner. In the training process, the labeled data are used to train the Diffuser for multi-task learning. \textcolor{blue}{Blue} arrows represent the planning process of the proposed DSC-Diffuser in horizon $H$, while \textcolor{red}{red} arrows represent the generation process in the Diffuser. Orange points is are the goals. \textcolor{green}{Green} arrows indicate the input of labeled trajectory data used to train the model during the training phase.}
    \label{fig-framework}
    \end{figure*} 

\section{Methodology}
\label{Methods}
Navigating signal-free intersections presents unique challenges in urban traffic due to the absence of traffic signals, requiring vehicles to adapt to dynamic and uncertain environments. This section details our problem formulation and the methodology used to generate human-like trajectories and ensure safe navigation at such intersections.

\subsection{Problem Formulation}
Our algorithm is specifically designed for navigating signal-free intersections, which are common in urban traffic environments and present unique challenges due to the absence of traffic signals. Unlike scenarios such as lane changing on a highway, in this context, the location of the target position after crossing is known. The problem is formulated as the conditional-generating problem. Specifically, suppose $\mathcal{F}(\cdot)$ is our model, which generates the human-like trajectories using the initial and end positions, along with the task as conditions. 

\begin{equation}
\label{prob}
\boldsymbol{\tau} = \mathcal{F}\left( s_t,a_t,s_{t+H}, \mathcal{M}, S_{obs}(t)\right) 
\end{equation}

\noindent where the trajectory $\boldsymbol{\tau}$ consists of a sequence of states and actions:

\begin{equation}
\label{state}
\boldsymbol{\tau}:=(s_t,a_t,s_{t+1},a_{t+1},...,s_{t+H},a_{t+H})
\end{equation}
where $t$ denotes the time at which a state was visited in trajectory, and the state is defined as $\boldsymbol{s_t}:=(x_t,y_t,v_{x_t},v_{y_t})$. The action at time $t$ is defined as $\boldsymbol{a_t}:=(a_{x_t},a_{y_t})$. $H$ denotes the trajectory horizon, and $s_{t+H}$ denotes the end position regarded as the planning goal.
The task $\mathcal{M}$, which can take one of three values, corresponds to different maneuvers: $\{Turning Left, Going Straight, Turning Right\}$, represented by $-1$, $0$, and $1$, respectively.
$S_{obs}(t)$  is the set of states of other vehicles which are regarded as obstacles for the ego vehicle and $S_{obs}(t):=\{s_{obs}^1(t), s_{obs}^2(t),...,s_{obs}^N(t)\}$. $N$ is the number of surrounding vehicles within the sensory range set as $8$ meters \cite{vivacqua2017low}. In each  planning step $t^\prime$, the state of the ego vehicle is updated by:
\begin{equation}
s_{t^\prime},a_{t^\prime}=\mathcal{F}\left(O_{t^\prime}, \tau_{t^\prime-1}|s_{t+H},\mathcal{M} \right)
\end{equation}
where $t^{\prime}$ is the planning time, and $t< t^{\prime}< t+H$. In this study, task-guided and goal-oriented diffusion models are used to recover policies from human driver datasets at signal-free intersections. Subsequently, the output of the diffusion models is used as a reference control, and DHOCBF is incorporated as a hard constraint to ensure driving safety, as shown in Fig. \ref{fig-framework}.

\subsection{Task-Guided and Goal-Oriented Planning with Diffuser}
At signal-free intersections, left-turn, right-turn, and through movements each require distinct trajectory planning strategies to account for their unique decision-making processes. To effectively accommodate multiple tasks within a unified planning framework, we incorporate classifier-free guidance, where task-specific conditions (encoded as $\boldsymbol{l}(\boldsymbol{\tau})$, representing maneuver-specific attributes) are explicitly integrated into the diffusion process. Therefore, integrating task labels into the trajectory generation process provides contextual information, enabling the model to effectively capture task-specific characteristics and generate trajectories that are smoother and more aligned with the intended maneuvers.

Diffusion model \cite{ho2020denoising} learns by reversing a controlled diffusion process, which consists of two main stages: a forward diffusion stage $q(\boldsymbol{\tau}^j|\boldsymbol{\tau}^{j-1})$ and a backward denoising stage $p_\theta(\boldsymbol{\tau}^{j-1}|\boldsymbol{\tau}^j,\boldsymbol{l}(\boldsymbol{\tau}))$.  Then the posterior distribution of the diffusion process from $\boldsymbol{\tau}^{0}$ to $\boldsymbol{\tau}^{N}$ is formulated as:

\begin{equation}
    q(\boldsymbol{\tau}_{1:N}|\boldsymbol{\tau} ^ 0) := \prod_{j=1}^{N}q\left(\boldsymbol{\tau}^{j}|\boldsymbol{\tau}^{j-1})\right)
\end{equation}

Correspondingly, the reverse diffusion process is given by:

\begin{equation}
    p(\boldsymbol{\tau}_{0:N}|\boldsymbol{l}(\boldsymbol{\tau})) := p(\boldsymbol{\tau}_N)\prod_{j=1}^{N}p_\theta \left(\boldsymbol{\tau}^{j-1}|\boldsymbol{\tau}^{j},\boldsymbol{l}(\boldsymbol{\tau})\right)
\end{equation}
where $p(\boldsymbol{\tau}_N)$ is an initial noise Gaussian distribution $x_{k-1}\sim\mathcal{N}(\mu_{k-1},\alpha\Sigma_{k-1})$.

Distinct from classifier guidance, the unconditional denoising diffusion model $p_\theta$ parameterized through a score estimator $\boldsymbol{\epsilon}_\theta(\boldsymbol{\tau}^j)$ is trained together with the conditional model $p_\theta(\boldsymbol{\tau}^j|\boldsymbol{l}(\boldsymbol{\tau}^i)) $ parameterized through $\boldsymbol{\epsilon}_\theta(\boldsymbol{\tau}^j,\boldsymbol{l}(\boldsymbol{\tau}))$. The score of classifier-free guidance \cite{ho2022classifier} is defined as:
\begin{equation}
\label{score}
\tilde{\boldsymbol{\epsilon}}=\boldsymbol{\epsilon}_\theta(\boldsymbol{\tau}^j,\emptyset,j)+w\boldsymbol{\epsilon}_\theta(\boldsymbol{\tau}^j,\boldsymbol{l}(\boldsymbol{\tau}),j)-w\boldsymbol{\epsilon}_\theta(\boldsymbol{\tau}^j,\emptyset,j) \end{equation}
where \(w\) is the guidance scale. In this work, there are two types of timesteps: one associated with the diffusion process and the other with the planning process. To maintain clarity, superscripts \(j \in \{1, \dots, N\}\) denote timesteps in the forward diffusion stage, while subscripts \(t \in \{1, \dots, H\}\) indicate the trajectory timesteps. When \(w = 0\), the conditional information has no impact on trajectory generation, whereas larger values of \(w\) can significantly strengthen this influence. Also, unreasonable results may be obtained if the guidance weight $w$ is too high. Then, the loss function can be defined as \cite{ho2022classifier}:

\begin{equation}
\label{loss function}
\begin{aligned}
\mathcal{L}(\theta)=&\mathbb{E}_{j,\boldsymbol{\tau},\epsilon}
\left[\left\|\epsilon-\epsilon_\theta\left(\boldsymbol{\tau}^j,(1-\beta)\boldsymbol{y}(\boldsymbol{\tau}^j)+\beta\emptyset,j\right)\right\|^2\right]\\
&+w_a\mathbb{E}_{a,\boldsymbol{\tau}^j}[||a-a'(\boldsymbol{\tau}^j)||^2
\end{aligned}
\end{equation}
where $\beta$  represents the probability of ignoring the conditional information, $\epsilon$ denotes the sample noise in the forward process, $a$ denotes the action set of the trajectory in the dataset, and $w_a$ denotes the weight of the action loss. The losses of the trajectories and action sets are summed and jointly used for optimization.

Vehicles are assumed to have predetermined lane assignments based on prior routing decisions before approaching an intersection. These predetermined arrival lanes serve as fixed goal locations for trajectory generation. To ensure feasible trajectories and minimize displacement errors, a goal-oriented constraint is imposed within the trajectory planning framework. Specifically, this constraint encourages that vehicles precisely reach their designated target lanes at the intersection exit. Integrating explicit goal constraints effectively reduces the accumulation of trajectory prediction errors, thereby enhancing trajectory accuracy and intersection navigation reliability. Referring to \cite{janner2022planning}, the constraint of the state is defined as:
\begin{equation}
\label{goal}
C(\boldsymbol{\tau})=\begin{cases}+\infty, &\text{if }\mathbf{g}_t=(\mathbf{s}_t,\mathbf{a}_t)\\\quad0, &\text{otherwise}\end{cases}
\end{equation}
where $\mathbf{g}_t$ is the goal set of state and action at time $t$. To implement this constraint, the sampled values are replaced by the goal $\mathbf{g}_t$ after all diffusion timesteps $i$.

\subsection{Safety-Critical Planning by DHOCBF}
\begin{table}[!ht]
    \centering
    \color{black}
    \caption{The Definitions of Acronyms
    \label{Acronyms}}
    \begin{tabular}{{ll}}
    \hline
        Acronym  & Definition \\ \hline
        BF & Barrier Function \\ 
        CBF &  Control Barrier Function \\ 
        HOCBF & High-Order Control Barrier Function  \\ 
        DHOCBF & Dynamic High-Order Control Barrier Function  \\ 
        DSC-Diffuser  & Dynamic Safety-Critical Diffuser \\ 
        ADE & Average Displacement Error \\ 
        FDE & Final Displacement Error \\ 
        SR & Success Rate  \\ 
        QP & Quadratic Program \\ \hline
    \end{tabular}
\end{table}

To enforce strict collision constraints on learned trajectory planning strategies, we integrate DHOCBF constraints as a safety-critical control mechanism to ensure the continuous generation of safe trajectories. The definitions of acronyms are shown in TABLE \ref{Acronyms}.  For the design of this controller, we consider an affine control system of the following form:
\begin{equation}
\label{affine system}
\dot{\boldsymbol{s}}_t=f(\boldsymbol{s_t})+g(\boldsymbol{s_t})\boldsymbol{u_t}
\end{equation}
where $\boldsymbol{s_t}\in\mathbb{R}^n$, $f{:}\mathbb{R}^n\to\mathbb{R}^n$,  $g{:}\mathbb{R}^n\to\mathbb{R}^{n\times q}$ are locally Lipschitz continuous, and $\boldsymbol{u}_t\in U\subset\mathbb{R}^q$ where $\textit{U}$ denotes a control constraint set. 

\textbf{Definition 1. (Set invariance)} A set $C\subset\mathbb{R}^n$ is forward invariant for system \ref{affine system} if its solutions for some $\boldsymbol{u}\in U$ starting at any ${\boldsymbol{s}_0}\in C$ satisfy $\boldsymbol{s_t}\in C$, $\forall t>0$. A safe set is defined as $ C = \{ \boldsymbol{s_t} \in \mathbb{R}^n \mid h(\boldsymbol{s_t}) \geq 0 \} $, where $ h: \mathbb{R}^n \to \mathbb{R} $ is a continuously differentiable function.

\textbf{Definition 2. (Barrier Function, BF)} The function $h$: $\mathbb{R}^n \to \mathbb{R}$ is a candidate BF for system \ref{affine system} if there exists a class $ \mathcal{K}$ function $\alpha$ such that:
\begin{equation}
\label{BF}
    \dot{h}(\boldsymbol{s})+\alpha(h(\boldsymbol{s}))\geq0, \quad\forall\boldsymbol{s}\in C.
\end{equation}

\textbf{Definition 3. (Control Barrier Function, CBF \cite{ames2014control})} A function $ h(\boldsymbol{s_t}) $ is a candidate control barrier function for a system if there exists an extended class $ \mathcal{K} $ function $ \alpha $ such that:
\begin{equation}
\label{eq-CBF}
    \sup_{\boldsymbol{u} \in U} \left[  L_{f}h(\boldsymbol{s_t}) + L_{g}h(\boldsymbol{s_t})\boldsymbol{u}\right] \geq -\alpha(h(\boldsymbol{s_t}))
\end{equation}  
where $ L_f $ and $ L_g $ represent the Lie derivatives along the functions $ f $ and $ g $, respectively. The set of all input $u$ that satisfy Eq.\ref{eq-CBF} for $\boldsymbol{s_t}$ can be defined as:
\begin{equation}
\label{Kcbf}
K_{cbf}(\boldsymbol{s_t})=\{\boldsymbol{u}\in\mathcal{U}:L_fh(\boldsymbol{s_t})+L_gh(\boldsymbol{s_t})\boldsymbol{u}+\alpha h(\boldsymbol{s_t})\geq0\}
\end{equation}

Then, a differential control system model for driving is defined:
\begin{equation}
\label{control system}
\left[ \begin{array}{c}
	\dot{x}_t\\
	\dot{y}_t\\
	\dot{v}_{x_t}\\
	\dot{v}_{y_t}\\
\end{array} \right] =\left[ \begin{array}{c}
	v_{x_t}\\
	v_{y_t}\\
	0\\
	0\\
\end{array} \right] +\left[ \begin{array}{c}
	0\\
	0\\
	1\\
	0\\
\end{array}\begin{array}{c}
	0\\
	0\\
	0\\
	1\\
\end{array} \right] \left[ \begin{array}{c}
	u_{x_t}\\
	u_{y_t}\\
\end{array} \right] 
\end{equation}

\begin{figure}[tbp] 
\centering 
  \includegraphics[width=1\linewidth]{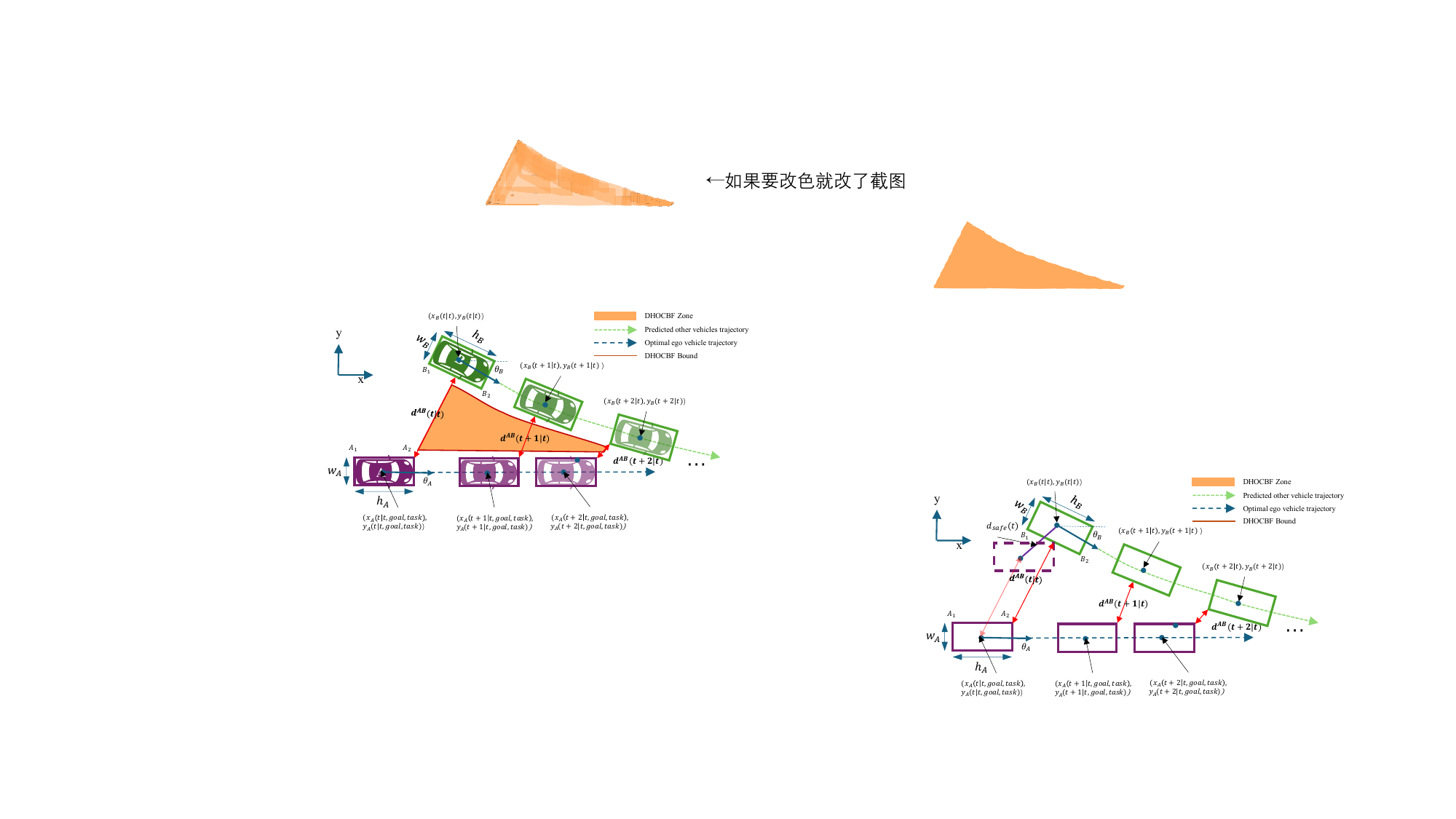}
  \caption{The optimal trajectory for ego vehicle A $p_A(t:t+H|t)$ to avoid a collision with other vehicle B $p_B(t:t+H|t)$, whose trajectory is predicted, at time $t$ over $H$ future steps. The safe distance, $d_{safe}(t)$ is dynamic. The HOCBF zone is the safe region, limited by HOCBF bounds, ensuring $h(\boldsymbol{s_t},\boldsymbol{s_{obs}(t)}) \geq 0$, so that the ego vehicle $A$ does not approach the other vehicle $B$ too closely.}
  \label{fig:subfig_1} 
\end{figure}

\noindent where $\boldsymbol{s_t}:=\left( x_t,y_t,v_{x_t},v_{y_t}  \right)$, $\boldsymbol{u}_t=\left( u_{x_t},u_{y_t} \right)$, $x_t,y_t$ denote the horizontal and vertical movement of vehicle, $v_{x_t},v_{y_t}$ denote the vertical and lateral speeds, $u_{x_t},u_{y_t}$ denote the control inputs for  vertical and lateral acceleration. Vertical and lateral directions are the directions of movement along the lane and perpendicular to the lane. The barrier function $h(\boldsymbol{s_t},\boldsymbol{s_{obs}^{i}\left( t \right)})=\left( x_t-x_{obs}^{i}\left( t \right) \right) ^2+\left( y_t-y_{obs}^{i}\left( t \right) \right) ^2-(d_{safe}^{i}(t))^{2}$, where $i=0,1,...,N-1$. In previous research, obstacles and vehicles are typically modeled as ellipses or circles \cite{10156163}. Given the limited space at signal-free intersections, representing vehicles as ellipses or circles can significantly lead to deadlock maneuvers \cite{thirugnanam2022safety}. Each vehicle can be represented as a rectangle, described by the position $\boldsymbol{p}^i\left( t \right) =\left( x^i\left( t \right), y^i\left( t \right) \right)$ and additional information $l^i\left( t \right) =\left( w^i,h^i,\theta ^i\left( t \right) \right) $, as shown in Fig.\ref{fig:subfig_1}.

$d_{safe}^{i}(t)$ is the minimum and dynamic safe distance between the other vehicle $i$ and ego vehicle at time $t$. Initially, the two closest points on the surfaces of the two vehicles are identified. The minimum distance between the corner of the rectangle and the surface of the other rectangle can be calculated by Eq.\ref{min dist}. The position of the point in the surface is determined by Eq.\ref{point}. Subsequently, the projection of the line segment formed by these points onto the centerline is calculated. Ultimately, the minimum safe distance is derived. As the vehicle travels approximately at the original angle for a brief period, the effect of angular acceleration on the minimum safe distance is neglected in the derivation; thus, 
\begin{equation}
\label{min dist}
d_{min}^{B_1A_1A_2}=\begin{cases}
	\left| \overrightarrow{A_1B_1} \right|,r\geq 1\\
	\left| \overrightarrow{A_2B_1} \right|,r\leq 0\\
	\left| \overrightarrow{A_1B_1} \right|\cdot sin\left( \left| \theta _A-\theta _B \right| \right), otherwise
\end{cases}
\end{equation}

\begin{equation}
\label{point}
p_{\text{surf}} = \left\{
\begin{aligned}
    & \quad\quad\quad\quad A_1, && r \geq 1 \\
    & \quad\quad\quad\quad A_2, && r \leq 0 \\
    & \left( x_{A_1} + r \cdot (x_{A_2} - x_{A_1}), \right. \\
    & \quad \left. y_{A_1} + r \cdot (y_{A_2} - y_{A_1}) \right), &&otherwise
\end{aligned}
\right.
\end{equation}

\noindent where $
r = \frac{\overrightarrow{A_1B_1}\cdot \overrightarrow{A_1A_2}}{\left| \overrightarrow{A_1A_2} \right|^2}
$.

Condition \ref{eq-CBF} ensures that if $ h(x) \geq 0 $, the system's state will remain within the safe set $C$ for all future times. Based on Eq.\ref{control system}, the CBF is calculated, and it is found that $L_{g}\boldsymbol{s_t}=0$, indicating that the control input $\boldsymbol{u}_t$ is unable to ensure system safety. Our control system exhibits 2nd-order dynamics, therefore standard CBFs may not be sufficient. High Order Control Barrier Functions (HOCBFs) \cite{xiao2019control} extend the concept of CBFs to include higher-order derivatives of the safety function $ h(x) $, thereby allowing for more robust safety guarantees. 

\begin{figure*}[tbp]
  \centering
  \begin{minipage}[c]{0.58\textwidth}  
    \centering
    \subfigure[MA]{
      \includegraphics[scale=0.4]{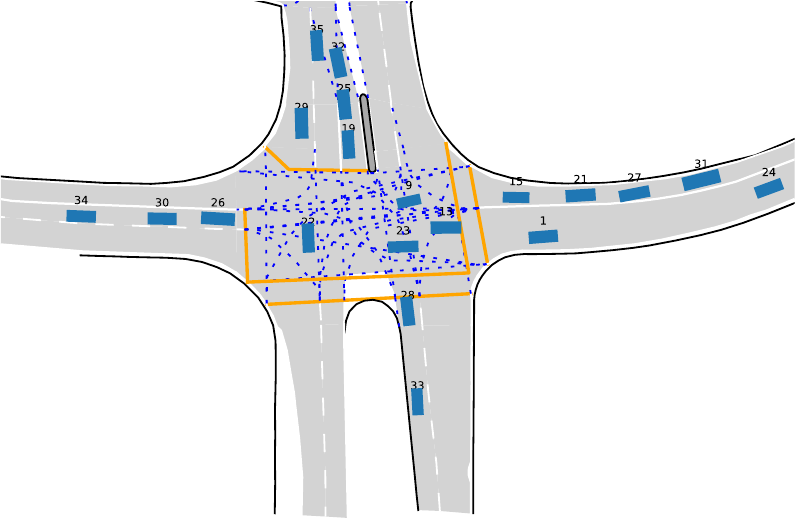}
    }
    \subfigure[GL]{
      \includegraphics[scale=0.4]{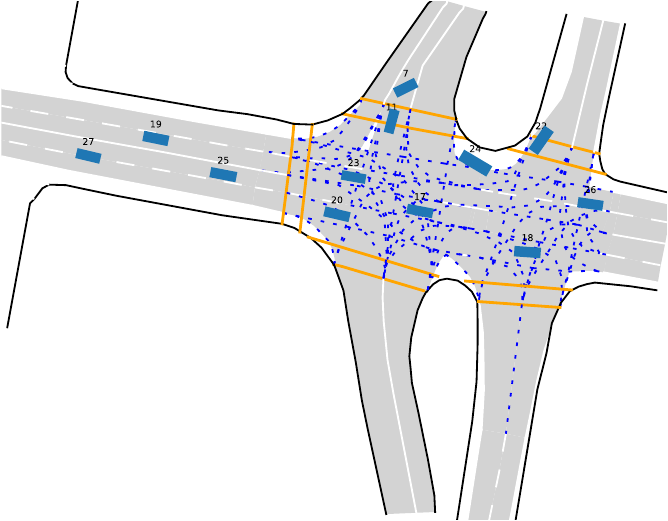}
    }
    \caption{MA and GL scenarios for given situations. The orange lines are stop lines and pedestrian lines.}
    \label{scenes}
  \end{minipage}%
\hspace{-4pt}
  \begin{minipage}[b]{0.4\textwidth}  
    \centering
    {\footnotesize
    \captionof{table}{The summary of the data}
    \label{info}
    \begin{tabular}{|c|c|c|}
        \hline
        Scenarios & Number of vehicles &  Video length (min) \\ \hline
        MA & 2982 & 107.37 \\ \hline
        GL & 10518 & 259.43 \\ \hline
    \end{tabular}
    }
  \end{minipage}

\end{figure*}

\textbf{Definition 4 (High Order Control Barrier Function (HOCBF) \cite{9516971})} A function $b$: $\mathbb{R}^n \times [t_0,\infty)\to\mathbb{R}$ is a candidate HOCBF for affine control system if it is $m$th-order differentiable and there exist differentiable class $ \mathcal{K} $ functions $\alpha_i$, $i\in\{1,\ldots m\}$ s.t.
\begin{equation}
\label{HOCBF_BF}
    \gamma_m(\boldsymbol{s_t})\geq0
\end{equation}
for all $(\boldsymbol{s_t})\in C_{1}(t)\cap,\ldots,\cap C_{m}(t)\times[t_{0},\infty).$ And, $\gamma_i(\boldsymbol{s_t})=\dot{\gamma}_{i-1}(\boldsymbol{s_t})+\alpha_i\gamma_{i-1}(\boldsymbol{s_t}),i\in\{1,\ldots,m\}$, $C_i(t)=\{\boldsymbol{s_t}\in\mathbb{R}^n:\gamma_{i-1}(\boldsymbol{s_t})\geq0\}$, $i\in\{1,\ldots,m\}.$ $\gamma_{0}(\boldsymbol{s_t})=h(\boldsymbol {x_t}).$ $\alpha_i$ denotes class $ \mathcal{K} $  functions. 

In dynamic CBF, $h(\boldsymbol{s_t},\boldsymbol{s_{obs}})$ is defined as a barrier function that incorporates information about the opponent's vehicle; thus
\begin{equation}
 \dot{h}(\boldsymbol{s_t},\boldsymbol{s_{obs}})= \frac{\partial h(\boldsymbol{s_t},\boldsymbol{s_{obs}(t)})}{\partial \boldsymbol{s_t}}\boldsymbol{\dot{s}_t}+
 \frac{\partial h(\boldsymbol{s_t},\boldsymbol{s_{obs}(t)})}{\partial \boldsymbol{s_{obs}(t)}}\boldsymbol{\dot{s}_{obs}(t)}
\end{equation}

By Nagumo's theorem, given a continuously differentiable constraint $h(\boldsymbol{s_t},\boldsymbol{s_{obs}(t)})\left( h(\boldsymbol{s_0},\boldsymbol{s_{obs}(t))} \geqslant 0 \right) $ and combining with \ref{HOCBF_BF}, the necessary and sufficient condition for guaranteeing the forward invariance of the safe set $C$ is:
\begin{equation}\gamma_m\left( \boldsymbol{s_t},\boldsymbol{s_{obs}(t)} \right) \geqslant 0, \forall \boldsymbol{s_t}\in C_1(t)\cap,\ldots,\cap C_m(t)
\end{equation}

$\alpha_1$ and $\alpha_2$ are linear and equal to $\beta_1$ and $\beta_2$, which are constants. 
In a dynamic environment, other vehicles are beyond our control, thus $u_{obs}=0$. It is assumed that all vehicles have the same control system model and information about the state of the opponent's vehicle can be obtained at each time step. Therefore, the 2nd-HOCBF constraint Eq.\cite{9516971} is reformulated for a dynamic environment as follows.
Here $h(\boldsymbol{s_t},\boldsymbol{s_{obs}^{i}\left( t \right)})$ is abbreviated to $h\left(\boldsymbol{s_t}\right)$.
\begin{equation}
\label{HOCBF}
\begin{aligned}-L_gL_fh(\boldsymbol{s_t})\boldsymbol{u}_{\boldsymbol{t}}\le &L_{f}^{2}h(\boldsymbol{s_t})+L_{f_{obs}}^{2}h\left( \boldsymbol{s_t} \right) \\&+
(\beta_1+\beta_2)\left( L_fh(\boldsymbol{s_t})+L_{f_{obs}}h(\boldsymbol{s_t}) \right) \\&+(\dot{\beta}_1+\beta_1\beta_2)h(\boldsymbol{s_t})
\end{aligned}
\end{equation}
where \begin{equation}
    \begin{aligned}
    \label{Lg}
L_gL_fh(\boldsymbol{s_t})& = [2(x_t-x_t^o),2(y_t-y_t^o)], \\
L_f^2h(\boldsymbol{s_t})& =2v_t^2, \\
L_fh(\boldsymbol{s_t})& =2(x_t-x_t^o)v_{x_t}+2(y_t-y_t^o)v_{y_t},\\
{L_f}_{obs}^2h(\boldsymbol{s_t})& = 2v^2_{obs}(t), \\
{L_f}_{obs}h(\boldsymbol{s_t})&=-2(x_t-x_{t}^{o})v_{x_t}^{o}-2(y_t-y_{t}^{o})v_{y_t}^{o}.
\end{aligned}
\end{equation}

In the given equations, $\beta_{1}$, $\beta_{2}$ are the adjustable parameters. The value of $\dot{\beta}_1\left(\boldsymbol{z}\right) $ is fixed at 0. Combined with the objective function, the safety controller can be modeled:

\begin{align}
\label{Function}
&\boldsymbol{u_t}^*\boldsymbol{=}argmin \left\|\boldsymbol{{u_t-u_t}^{ref}}\right\|^2 \nonumber \\
s.t.
-L_g&L_fh^i\left( \boldsymbol{s_t} \right) \boldsymbol{u}_t\le L_{f}^{2}h^i\left( \boldsymbol{s_t} \right) +L_{f_{obs}}^{2}h^i\left( \boldsymbol{s_t} \right) \nonumber\\&
+\left( \beta _1+\beta _2 \right)
\left( L_fh^i\left( \boldsymbol{s_t} \right) +L_{f_{obs}}h^i\left( \boldsymbol{s_t} \right) \right) 
\nonumber\\&
+\left( \beta _1\beta _2\right) h^i\left( \boldsymbol{s_t} \right),  i\boldsymbol{=}0,...,N\boldsymbol{-}1\nonumber\\ 
&\begin{aligned}[t]
    \boldsymbol{u}_{\min}\leq \boldsymbol{u}_{\boldsymbol{t}}&\leq \boldsymbol{u}_{\max}\\
\boldsymbol{s_t}^i\in\mathcal{X}, \boldsymbol{u}_t&\in\mathcal{U}, t=0,...,H-1
\end{aligned}
\end{align}
where $\boldsymbol{{u_t}^{ref}}$ denotes the reference control of trajectories, which is calculated by the output of the planner $\boldsymbol{u(t)}$ and current state $\boldsymbol{s_t}$ to make our safe trajectories close to the reference trajectories of diffuser-based planner, and $\mathcal{X}$, $\mathcal{U}$ are the sets of admissible states and inputs, respectively. Using the following quadratic program (QP)-based controller design, the input $\boldsymbol{{u_t}^{ref}}$ given by the global planner can be minimally modified by the input $\boldsymbol{u^*}$ that satisfies Eq.\ref{HOCBF}. It is important to note that the safety controller acts more like a safety modifier.

\section{Experiments}
\label{Experiments}
The experimental settings, dataset, and data processing are first described. Subsequently, the assessment metrics and comparison algorithms are presented. A series of experiments in the same and a different signal-free intersections, distinct from the training scenario, are conducted to demonstrate the effectiveness and generalization of our approach by comparing it with other algorithms, including rule-based and imitation learning methods. 

\subsection{Datasets}

Our framework is evaluated using the Interaction dataset \cite{interactiondataset}, which provides naturalistic motion data from diverse traffic participants across various highly interactive driving scenarios in different countries. Data from two unsignalized intersection scenarios in the USA (GL and MA) are used, as shown in Fig.\ref{scenes} and TABLE \ref{info}. This study addresses the trajectory planning problem for crossing signal-free intersections while disregarding the stop-line crossing rule. Thus, trajectories between the point of crossing the pedestrian line and the pedestrian line at the target intersection are intercepted. All models are trained in the MA scene, which includes approximately 230 trajectories for each movement: turning right, going straight, and turning left. 

Two experiments are conducted: one in the MA scene, the same as the training scene, and another in the GL scene, which is distinct. To showcase the multi-task learning ability and safety guarantees of methods, training and testing on the MA scenario are conducted. In contrast, the GL scenario was left untrained, serving as a unseen scene to assess the generalization capabilities and safety guarantees of algorithms. Notably, all algorithms use the same processed, collision-free trajectory data.

\subsection{Comparison Algorithms}

 \begin{table}[tbp]
    \centering
    \color{black}
    \caption{The Value of Hyperparameters
    \label{params}}
    \begin{tabular}{ll}
    \hline
        Hyperparameter & Value \\ \hline
        UNet Dimensions & 32 ,64, 128, 256 \\ 
        Kernel Size  & 5 \\ 
        Condition Dropout & 0.1 \\ 
        Diffusion Timesteps & 20 \\ 
        Action Weight $w_a$ & 10 \\ 
        Loss Discount & 1 \\ \hline
    \end{tabular}
\end{table}

\begin{figure*}[t]
    \begin{minipage}[t]{0.48\textwidth}
        \centering
        \color{black}
        \subfigure[Distance to Obstacles Over Time for HOCBF and DHOCBF with Varying Obstacle Speeds]{
            \includegraphics[width=0.9\linewidth]{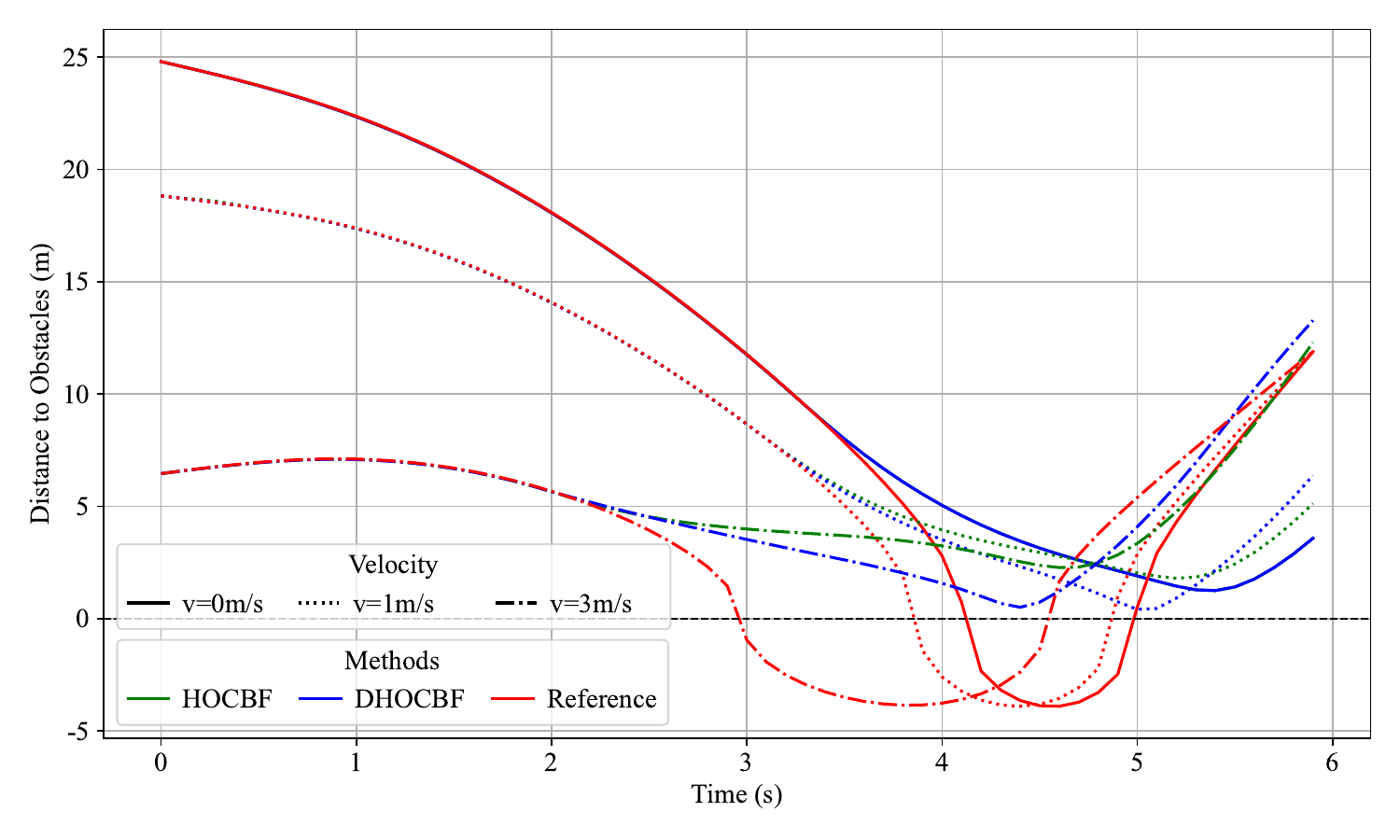}
            \label{fig:d2}}
        
        \subfigure[Generated Trajectories of HOCBF and DHOCBF with Varying Obstacle Speeds]{
            \includegraphics[width=0.9\linewidth]{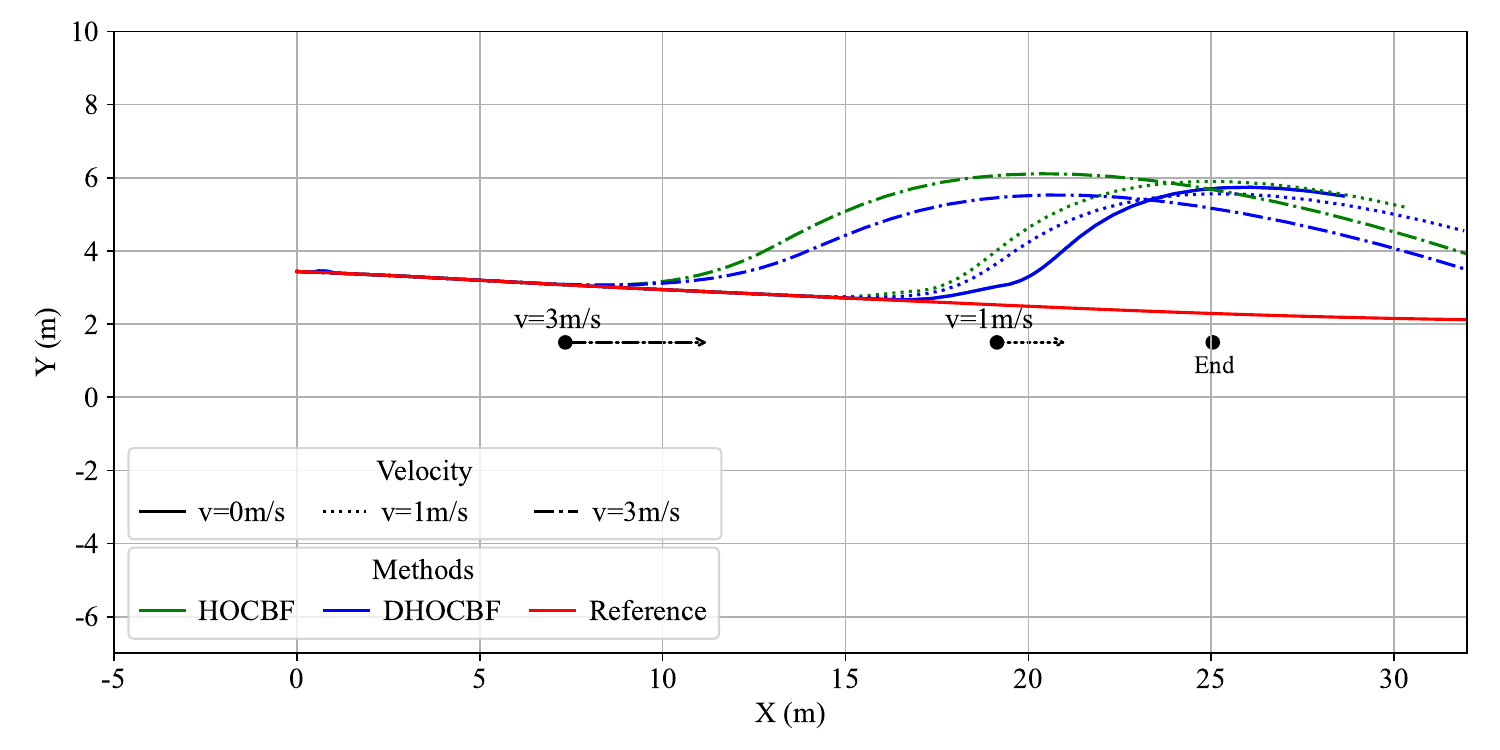}
            \label{fig:t2}}
        \caption{Performance Comparison of HOCBF and DHOCBF with Dynamic Obstacles of Varying Speeds. When the obstacle is stationary (v=$0$$m/s$), DHOCBF and HOCBF generate the same trajectories, resulting in the \textcolor{green}{green} and \textcolor{blue}{blue} lines overlapping in the figure. The speed of obstacles is set as $0, 1$ $m/s$, $3$ $m/s$. In this case, the ego vehicle is traveling in the same direction.}
        \label{fig:V}
    \end{minipage}%
    \hspace{0.04\textwidth} 
    \begin{minipage}[t]{0.48\textwidth}
        \centering
        \color{black}
        \subfigure[Distance to Obstacles Over Time for HOCBF and DHOCBF with Varying Radii]{
            \includegraphics[width=0.9\linewidth]{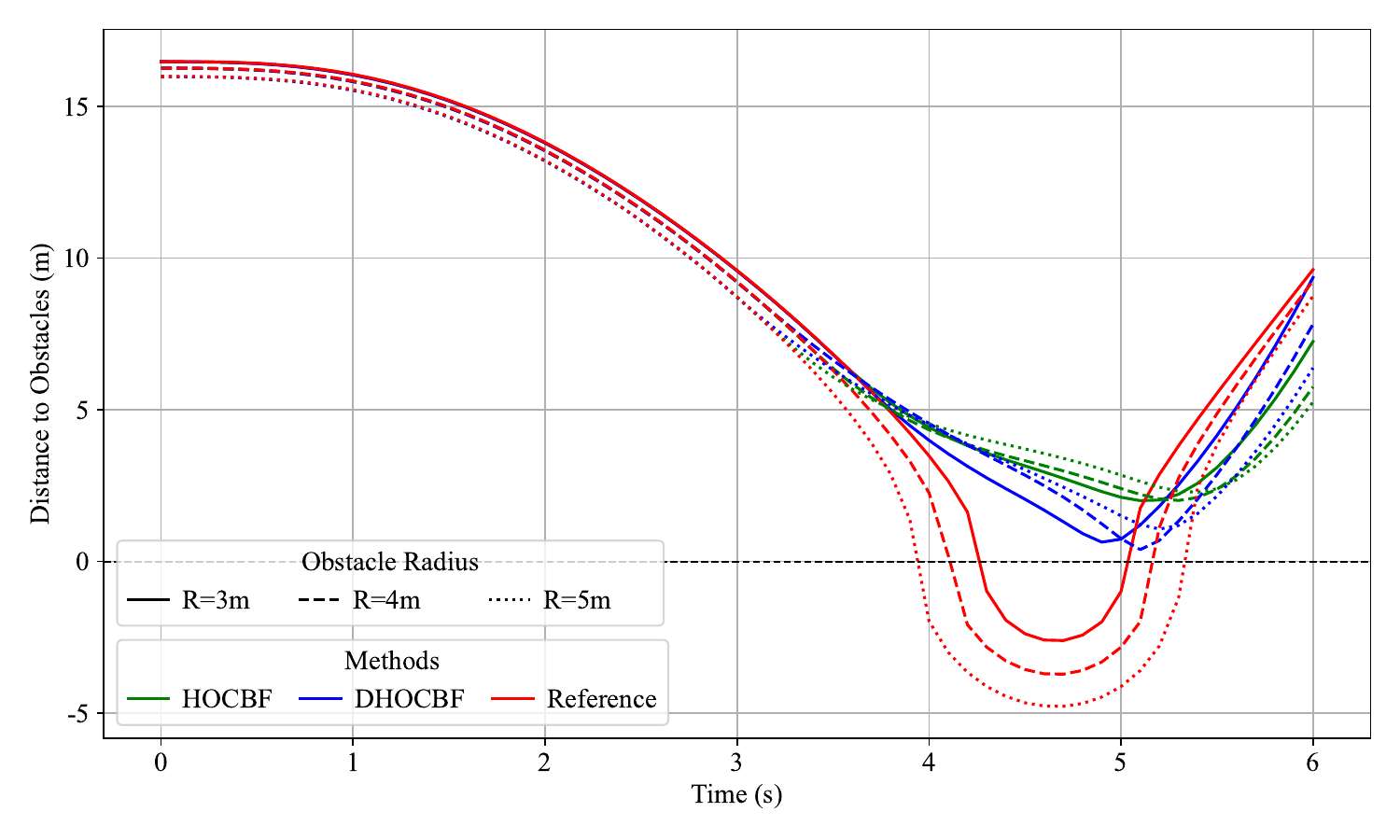}
                \label{fig:d1}}
        \
        \subfigure[Generated Trajectories of HOCBF and DHOCBF with Varying Obstacle Radii]{
            \includegraphics[width=0.93\linewidth, trim=0cm 0cm 0cm 0.4cm]{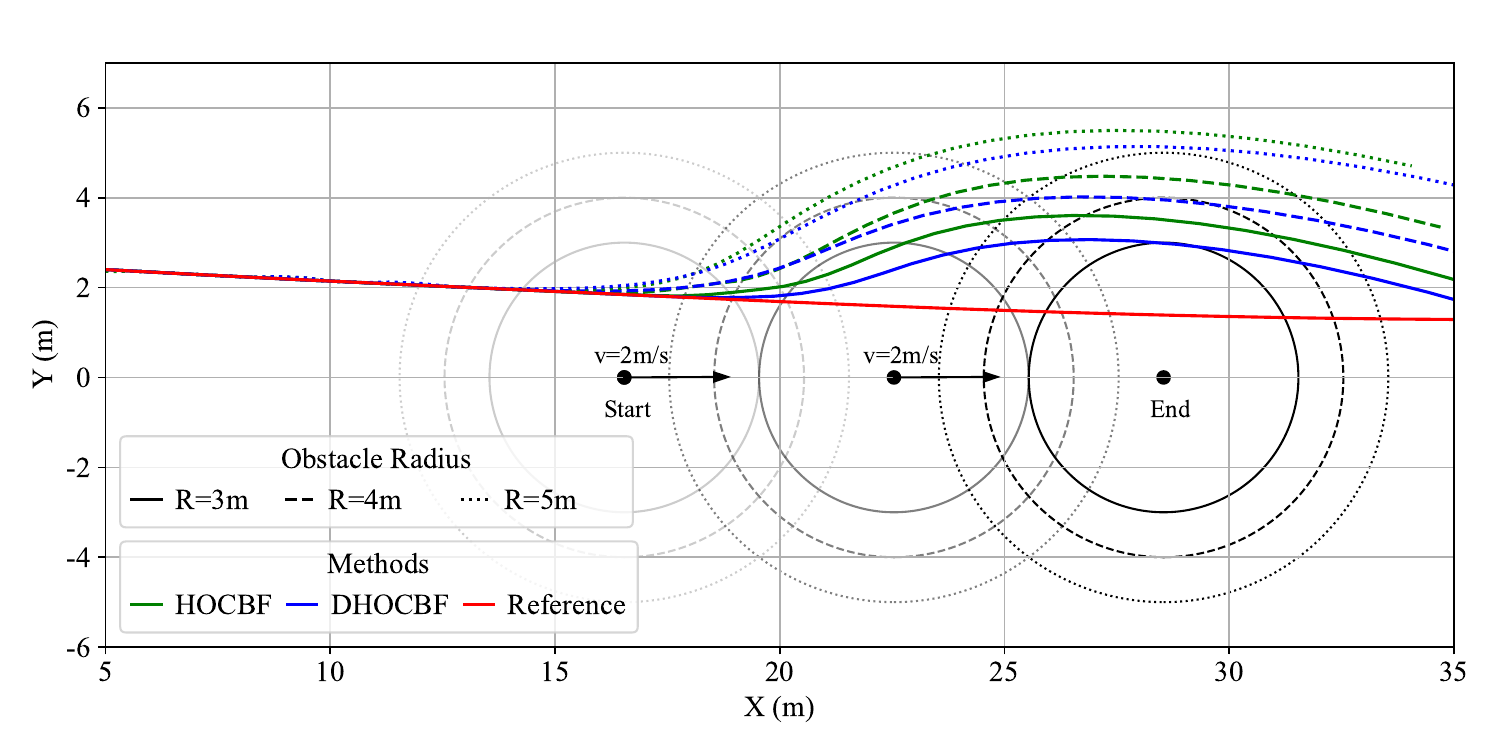}
            \label{fig:t1}}
        \caption{Comparative Analysis of HOCBF and DHOCBF with Varying Obstacle Radii. The obstacles move at a speed of $2$ $m/s$ in the positive x-direction.}
        \label{fig:R}
    \end{minipage}
\end{figure*}

\subsubsection{Human driver} The data from Interaction dataset \cite{interactiondataset}.
\subsubsection{IDM} The Intelligent Driver Model (IDM) calculates acceleration based on the current speed and the gap to the leading vehicle. Referring to \cite{jamgochian2023shail}, the closest vehicle that lies within 8 meters of the ego’s planned path and with a heading difference of less than $15^{\degree}$ is chosen as the leader vehicle. This model poses challenges in these driving scenarios, where it is unclear which vehicle the ego should "follow". A desired speed of $9.63$ m/s, a minimum spacing of $2.5$ m, a desired time headway of $1.6 s$, a nominal acceleration of $2.0$ $m/s^2$, and a comfortable braking deceleration of $3.0$ $m/s^2$ are set. These settings may vary across different scenarios \cite{wiseman2022autonomous}. In this study, all parameter values are referenced to the Interaction datasets \cite{interactiondataset} and \cite{albeaik2022limitations}.
\subsubsection{BC} Behavior Cloning (BC) \cite{torabi2018behavioral} method mimics a subset of the dataset via supervised learning, which is implemented to regress features to a mean and standard deviation that parameterize a normal distribution for ego vehicle acceleration. The model was trained by minimizing the negative log-likelihood of expert actions.
\subsubsection{GAIL} Generative Adversarial Imitation Learning (GAIL) aims to infer the latent reward function from the dataset and derive the driving policy by optimizing the learned reward. The agent is trained using the optimization objective and PPO as described in \cite{ho2016generative}.
\subsubsection{SHAIL} Safety-Aware Hierarchical Adversarial Imitation Learning (SHAIL) \cite{jamgochian2023shail} builds on GAIL by introducing high-level option selection to enhance safety. Details of this algorithm can be found in \cite{jamgochian2023shail}.
\subsubsection{GameFormer} GameFormer \cite{huang2023gameformer} is a Transformer-based prediction and planning framework that leverages level-$k$ game theory.
\subsubsection{DIPP} Differentiable Integrated Prediction and Planning (DIPP) \cite{10154577} method is a differentiable structured learning framework that utilizes a Transformer-based model as the predictor and a differentiable nonlinear optimizer as the motion planner.
\subsubsection{DSC-Diffuser} The proposed DSC-Diffuser method is used to learn policies from datasets and maintain safety. As described in Section III.A, we conduct conditional trajectory generation on the dataset, where the trajectory endpoints serve as GOALs, and left turns, right turns, and through movements define the TASK. Additionally, we apply DHOCBF as a safety modifier to ensure the security of the generated trajectories. To explore the roles of task guidance, goal orientation, and DHOCBF, ablation experiments are conducted. The value of parameters of DSC-Diffuser is shown in TABLE \ref{params}.

\subsection{Metrics}
Average Displacement Error (ADE), Final Displacement Error (FDE), and success rate (SR) \cite{10440492} are used to demonstrate the realism of our approach in comparison to human drivers and to assess the safety of these planning methods. The SR is defined as the ratio of the number of non-colliding trajectories to the total number of trajectories. Since the positions of the last step as the goals are used in the DSC-diffuser, the FDE of the algorithm’s output is fixed at zero. Therefore, the displacement error of the penultimate step is used as the FDE for the DSC-diffuser.

\section{Results And Discussion}
\label{Results}

In this section, the effectiveness of DHOCBF is first validated in maintaining safety within dynamic environments by comparing it with HOCBF under various conditions. Then, to further evaluate the effectiveness of the proposed DSC-Diffuser method, a comparison was conducted with other baseline methods in the same scenarios. Additionally, different guidance strengths, as well as the use of DHOCBF and goal setting, were tested to demonstrate the role of each module. All experimental results were obtained from five independent trials, each using a different random seed, to ensure statistical evidence. The average values from the tests in the MA and GL are shown in TABLE \ref{result_MA} and TABLE \ref{result_GL}, respectively.

\begin{figure}[t]
    \centering
    \includegraphics[width=0.92\linewidth]{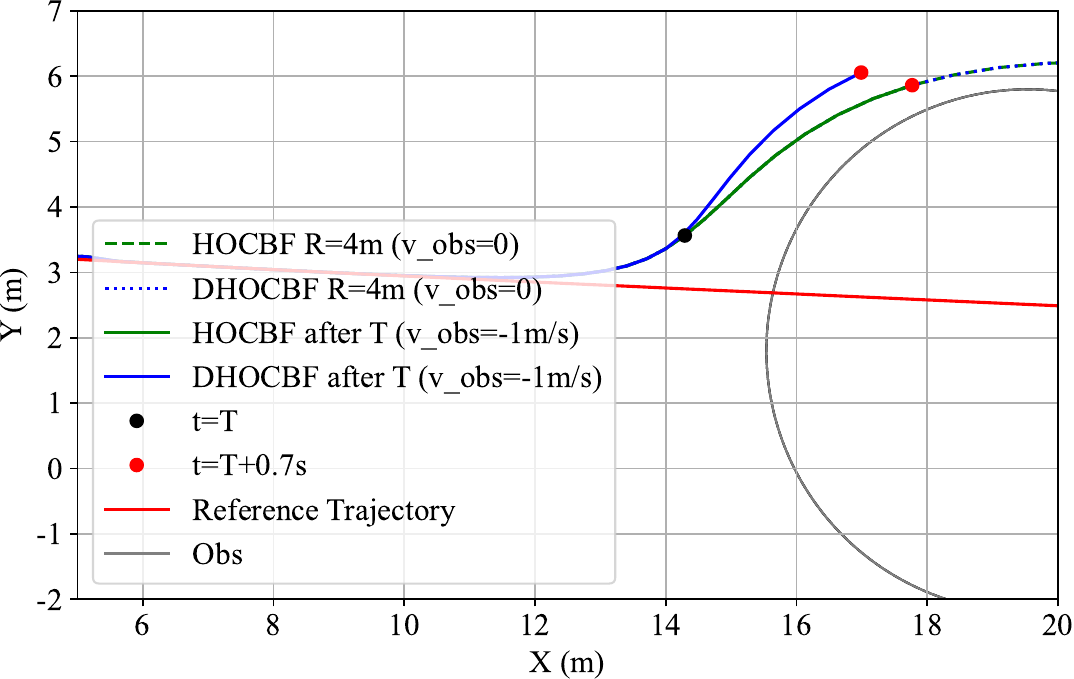}
    \caption{Generated Trajectories of HOCBF and DHOCBF under Environmental Disturbance. Initially, the obstacle is stationary, and at time $T$, it is assigned a velocity of $v=-1$ $m/s$, representing the uncertainty from the environment. The position of the obstacle here is the position at time $T$.}
    \label{fig:DS}
\end{figure}

\begin{figure}[ht]
    \centering
    \subfigure[Distance to Obstacles Over Time]{
        \includegraphics[width=0.9\linewidth]{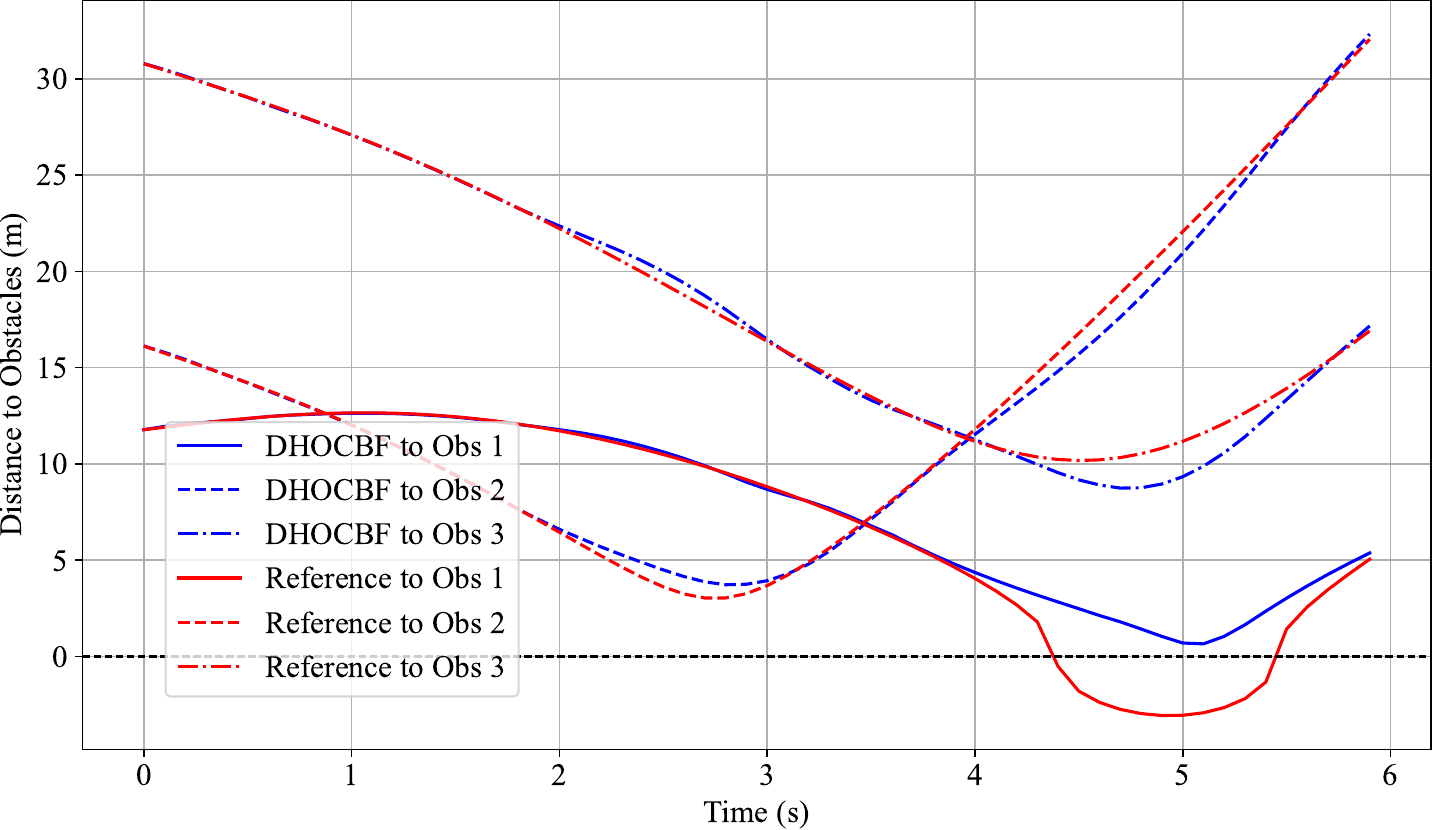}
        \label{fig:d3}}
    
    
    \subfigure[Generated Trajectories of DHOCBF in Complex Scenarios]{
        \includegraphics[width=0.93\linewidth]{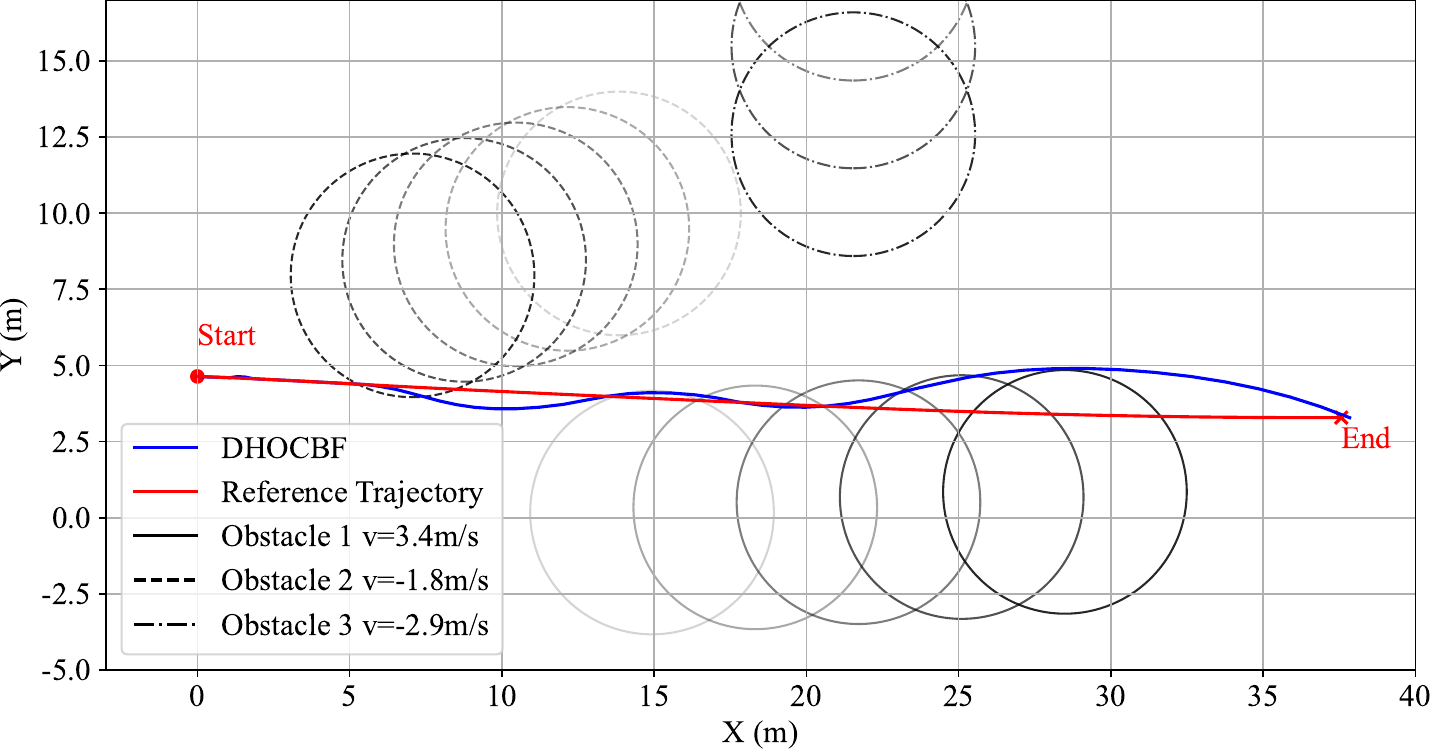}
        \label{fig:t3}}
    \caption{Distance to Obstacles and Generated Trajectories of the DHOCBF in Complex Scenarios}
    \label{fig:Multi-obs}
\end{figure}

\subsection{Validity Experiment of DHOCBF}
To demonstrate the adaptability of the proposed DHOCBF in dynamic environments, its ability to ensure safety is verified in scenarios with different obstacle velocities (Fig.\ref{fig:V}), varying obstacle sizes (Fig.\ref{fig:R}), environmental perturbations (Fig.\ref{fig:DS}), and multiple obstacles (Fig.\ref{fig:Multi-obs}), by comparing it with HOCBF. For clearer visualization, obstacles are simplified to circles, and the size of the ego vehicle is disregarded, so any distance greater than $0$ is considered safe. Infeasible reference trajectories are provided, allowing the distance to obstacles to reflect the safety and conservatism of the two constraints, HOCBF and DHOCBF. The distances between the ego vehicle and obstacles are shown in Fig.\ref{fig:d2}, Fig.\ref{fig:d1} and Fig.\ref{fig:d3}. The trajectories of reference and HOCBF-generated and DHOCBF-generated are shown in Fig.\ref{fig:t2}, Fig.\ref{fig:t1}, Fig.\ref{fig:DS} and Fig.\ref{fig:t3}.

The direction of motion of the original trajectory is from left to right. The original trajectory is nearly aligned with the positive $X$ direction, with speeds consistently greater than $5.2$ $m/s$. Additionally, to represent the movement of an obstacle, multiple circles are used, where lighter colors indicate earlier positions over time. Both the obstacles and the ego vehicle start moving from time $0$, except in Fig.\ref{fig:DS}. The optimal values $\beta_1$ and $\beta_2$ of $K$ functions are found by repeated experiments such that the quadratic programs(QPs) are feasible and the barrier function $h(x)$ is minimized when the constraints first become active. The optimal value varies under various conditions of HOCBF and DHOCBF.

\begin{table*}[b]
\centering
\color{black}
\caption{The Values of Comparison Metrics in MA Scene}
\label{result_MA}
\renewcommand\arraystretch{1.2}
\resizebox{0.8\linewidth}{!}{
\tiny
\begin{tabular}{ccccccc}
\hline
\multicolumn{7}{c}{MA}                                                                                                                                                                                                              \\ \hline
& Guidance & ADE      & VAR\_ADE & FDE      & VAR\_FDE & SR\\ \hline
BC \cite{torabi2018behavioral}                                & -        & 1.7472   & 0.1719   & 6.3376   & 0.4763   & 0.7879 \\ 
GAIL \cite{ho2016generative}                              & -        & 1.6622   & 0.3458   & 4.6168   & 0.7356  & 0.8650  \\ 
SHAIL  \cite{jamgochian2023shail}                            & -        & 2.9260  & 0.3542   & 4.0916   & 0.5134   & 0.8989 \\ 
GameFormer \cite{huang2023gameformer}                                & -      & 2.1739  & 0.0390       & 5.7785  & 0.1707   & 0.9149      \\ 
DIPP \cite{10154577}                                & -       & 3.7194 & 1.855E-04   & 4.1858  & 3.092E-04     & 0.9063      \\ 
IDM \cite{treiber2000congested}                                & -        & 3.4513  & 0        & 9.7487  & 0        & -      \\ 
Diffuser w/o Goals & 8 & 1.0894  & 0.0177  & 2.6913  & 0.1292  & \textbf{1}  \\ 
Diffuser w/  Goals & 0 & \textbf{1.756E-03} & 6.914E-10
 & \textbf{2.362E-03} & 2.797E-08 & \textbf{1} \\ 
DSC-Diffuser w/o Goals & 8 & 1.0869   & 0.0175   & 2.7043   & 0.1275   & \textbf{1} \\ 
DSC-Diffuser (ours) & 0 & 1.987E-03 & \textbf{3.558E-10} & 3.024E-03 & \textbf{2.001E-08} & \textbf{1} \\ \hline
\end{tabular}
}
\end{table*}

\begin{enumerate}
    \item Fig.\ref{fig:d2}, Fig.\ref{fig:d1} and Fig.\ref{fig:d3} show that both DHOCBF-generated and HOCBF-generated trajectories are safe when the ego vehicle and obstacles travel in the same direction under different conditions, as the distances to obstacles are consistently greater than $0$. These results demonstrate that trajectories generated by HOCBF and DHOCBF can adapt to variations in obstacle velocity and size.
    
    \item When comparing the trajectories in Fig.\ref{fig:d2} and Fig.\ref{fig:t2}, the gap between the HOCBF-generated trajectory and the reference trajectory increases with obstacle speed, highlighting HOCBF's conservatism in dynamic environments. The DHOCBF-generated trajectories more closely follow the reference trajectory, while maintaining smaller safe distances to the obstacle and demonstrating DHOCBF's adaptability to dynamic environments, avoiding overly conservative behavior.
    
    \item In Fig.\ref{fig:DS}, the response of two constraints to environmental perturbations is demonstrated. When the obstacle is stationary without perturbations, both HOCBF and DHOCBF produce the same trajectories. At the time $T$, after the obstacle’s velocity changes to $-1$ $m/s$ (opposite to the ego vehicle’s direction of driving) from 0, the DHOCBF-generated trajectory adjusts to maintain a greater distance from the obstacle, whereas the HOCBF-generated trajectory remains unchanged as if the obstacle were stationary. DHOCBF demonstrates the ability to maintain driving safety in uncertain environments, exhibiting robustness to perturbations.
    
    \item DHOCBF is robust in a complex environment, as shown in Fig.\ref{fig:Multi-obs}, which presents the safe trajectory navigating 3 surrounding objects. 
    
\end{enumerate}


\subsection{Performance Evaluation in MA Scene}
Among the baseline methods in the MA scenario, BC demonstrates relatively high error rates, with an ADE of 1.7472 and a FDE of 6.3376. This result points to BC’s limitations in effectively generalizing learned behaviors. GAIL significantly reduces these errors, achieving an ADE of 1.6622 and an FDE of 4.6168, reflecting the benefits of adversarial training. However, GAIL exhibits relatively high variability (VAR\_ADE of 0.3458 and VAR\_FDE of 0.7356), suggesting some inconsistency in predictions. SHAIL achieves a high SR of 0.8989; however, its ADE of 2.9260 is notably higher compared to GAIL, although its FDE of 4.0916 slightly improves upon GAIL. This mixed performance indicates that while SHAIL is more stable in terms of success rate, it does not substantially outperform GAIL in terms of displacement error.

GameFormer improves the success rate further (SR of 0.9149) but shows an increased ADE of 2.1739 and an FDE of 5.7785, implying a trade-off between stability and accuracy. DIPP exhibits higher errors (ADE of 3.7194, FDE of 4.1858), reflecting limitations in accurately capturing complex agent interactions. IDM, a rule-based model, shows significantly higher errors (ADE of 3.4513, FDE of 9.7487) and has zero variance (VAR\_ADE and VAR\_FDE equal to 0) precisely because of its deterministic, rule-based nature, highlighting its inadequacy for capturing dynamic complexity inherent in multi-agent scenarios.

In comparison, Diffuser models substantially outperform the baseline methods. Diffuser without goals (with a guidance weight of 8) achieves significantly reduced errors (ADE of 1.0894 and FDE of 2.6913) along with notably lower variance, demonstrating robust multi-task learning capabilities. Introducing explicit goals further enhances performance dramatically, reducing ADE and FDE to exceptionally low values (1.756E-03 and 2.362E-03, respectively) while maintaining a success rate of 1.

The DSC-Diffuser models exhibit similarly impressive performance. DSC-Diffuser without goals closely matches the Diffuser without goals, achieving an ADE of 1.0869 and an FDE of 2.7043, underscoring the effectiveness of the diffusion process. However, the complete DSC-Diffuser model (ADE of 1.987E-03 and FDE of 3.024E-03) performs slightly worse than the complete Diffuser model (ADE of 1.756E-03 and FDE of 2.362E-03). This small discrepancy suggests that while the integration of HOCBF significantly enhances safety, it introduces some conservatism due to fixed parameters, marginally limiting trajectory prediction accuracy compared to the Diffuser. Nonetheless, DSC-Diffuser maintains extremely low errors and the lowest variance among the evaluated models, confirming its superior stability and robustness in trajectory planning.


\subsection{Performance Evaluation in GL Scene}
\begin{table*}[t]
\centering
\color{black}
\caption{The Values of Comparison Metrics in GL Scene}
\label{result_GL}
\renewcommand\arraystretch{1.2}
\resizebox{0.8\linewidth}{!}{
\tiny
\begin{tabular}{ccccccc}
\hline
\multicolumn{7}{c}{GL}                                                                                                                                                                                                           \\ \hline
           &Guidance & ADE & VAR\_ADE & FDE & VAR\_FDE & SR \\ \hline
        BC\cite{torabi2018behavioral}  & - & 11.7211  & 0.0302  & 19.2179  & 0.1472  & 0.7757  \\ 
        GAIL\cite{ho2016generative} & - & 9.6124  & 0.2864  & 13.5379  & 0.9558  & 0.7583  \\ 
        SHAIL\cite{jamgochian2023shail}   & - & 9.9487  & 1.2143  & 14.8689  & 1.1078  & 0.8050  \\ 
        GameFormer\cite{huang2023gameformer}  & - & 3.1155  & 0.0075  & 7.8051  & 0.0590  & 0.8371  \\ 
        DIPP\cite{10154577}  & - & 5.5012  & 1.166E-04  & 6.4552  & 1.578E-04  & 0.8927  \\ 
        IDM\cite{treiber2000congested} & - & 13.2094 & 0 & 21.7039 & 0 & - \\ 
        Diffuser w/o Goals & 19 & 1.8111  & 0.0029  & 4.5289  & 0.0224  & 0.8780  \\ 
        Diffuser  w/ Goals & 3 & \textbf{0.1333}  & \textbf{1.967E-08} & \textbf{0.0288}  & \textbf{1.664E-07} & 0.9756  \\ 
        DSC-Diffuser w/o Goals & 19  & 1.7458  & 0.0024  & 4.3286  & 0.0215  & \textbf{1} \\ 
        DSC-Diffuser (ours) & 8  & 0.1334  & 6.680E-08 & 0.0304  & 3.469E-07 & \textbf{1} \\ \hline
\end{tabular}
}
\end{table*}

Performance analysis in the GL (untrained) scenario provides deeper insights into the model’s generalization capabilities. Baseline methods demonstrate substantially higher displacement errors in the GL scenario than in the MA (trained) scenario, with BC and IDM exhibiting notably high errors (ADE of 11.7211 and 13.2094; FDE of 19.2179 and 21.7039, respectively). GAIL and SHAIL show moderate improvements relative to BC but still exhibit elevated ADE and FDE values, accompanied by high variance, indicating challenges in achieving stable trajectory planning in untrained environments.

GameFormer and DIPP demonstrate improved accuracy and generalization compared to other baseline methods in the GL scenario, with ADE values of 3.1155 and 5.5012 and FDE values of 7.8051 and 6.4552, respectively. However, their errors remain substantially higher than those of diffuser-based models, highlighting their limited ability to learn a multi-task policy.

Diffuser-based methods exhibit strong generalization capabilities, with the Diffuser model without explicit goals achieving significantly lower errors (ADE of 1.8111, FDE of 4.5289). Incorporating explicit goal conditioning further significantly improves performance, yielding exceptionally low displacement errors (ADE of 0.1333, FDE of 0.0288) and minimal variance. DSC-Diffuser closely matches the exceptional performance of the Diffuser (ADE of 0.1334, FDE of 0.0304) while effectively ensuring safety (SR = 1). The marginally higher errors compared to Diffuser underscore DSC-Diffuser’s capability in preserving safety and ensuring trajectory feasibility, trading off only a minimal amount of accuracy to achieve robust performance in trajectory planning in unseen scenarios.
\subsection{Hyperparameter Sensitivity Analysis}
\begin{figure}[t]
    \centering
    \subfigure[Diffuser w/o Goals]{
        \includegraphics[width=0.9\linewidth]{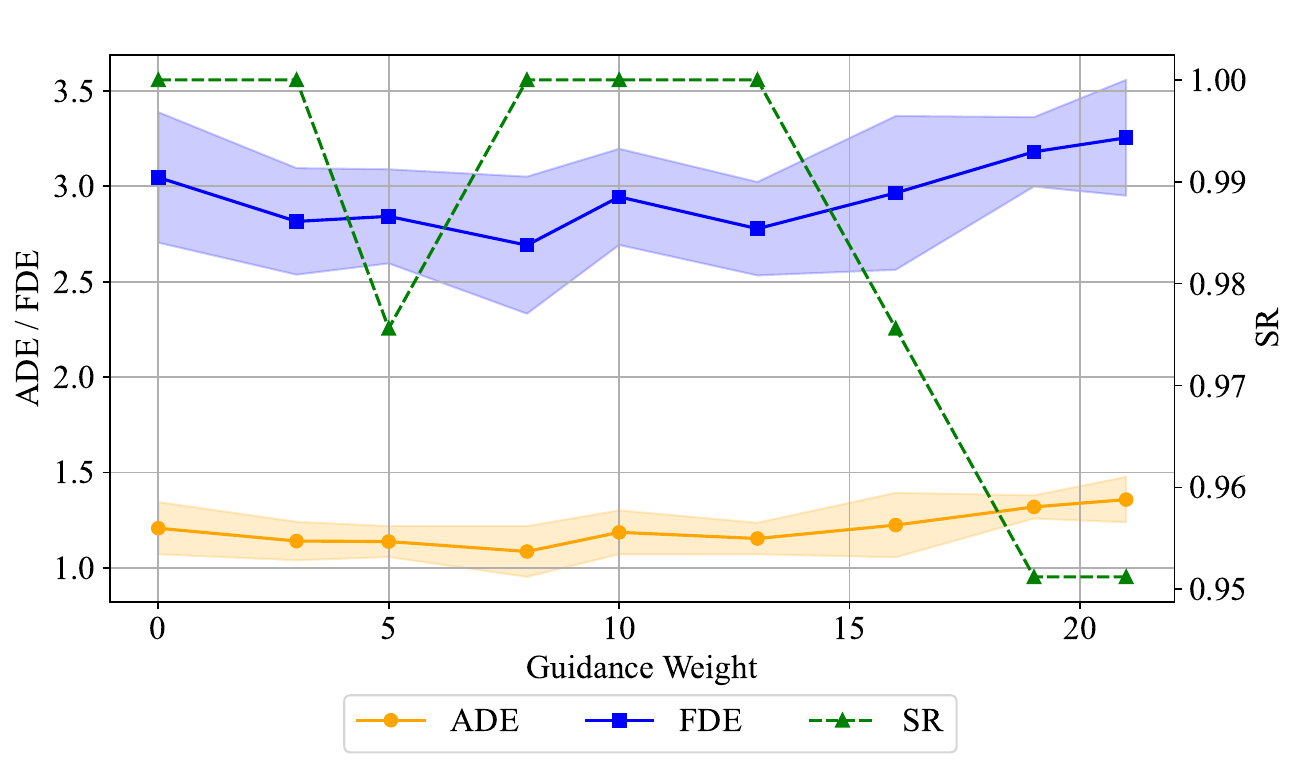}
        \label{fig:MA_w/o}}
    
    
    \subfigure[Diffuser w/ Goals]{
        \includegraphics[width=0.93\linewidth]{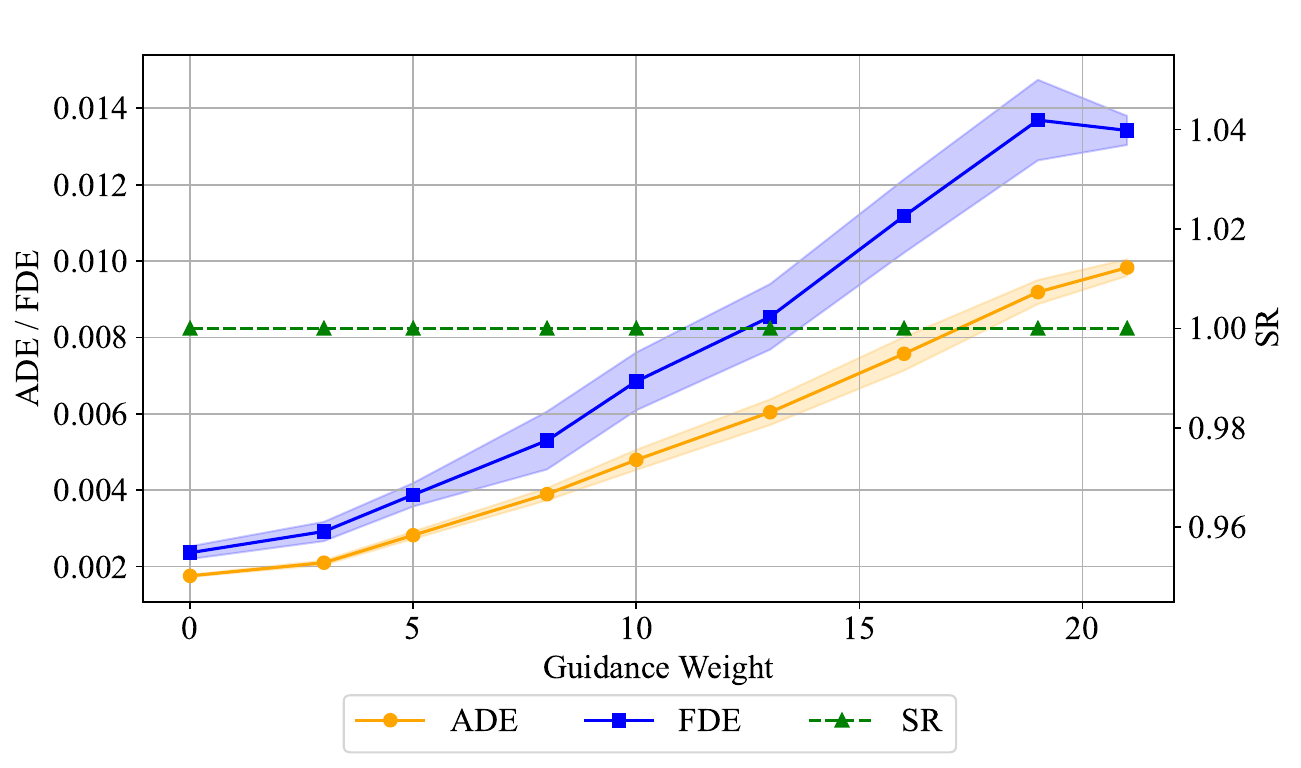}
        \label{fig:MA_w}}
    \caption{Impact of Guidance Weight on Model Performance in MA}
    \label{fig:MA}
\end{figure}

\begin{figure}[t]
    \centering
    \subfigure[\textcolor{black}{Diffuser w/o Goals}]{
        \includegraphics[width=0.9\linewidth]{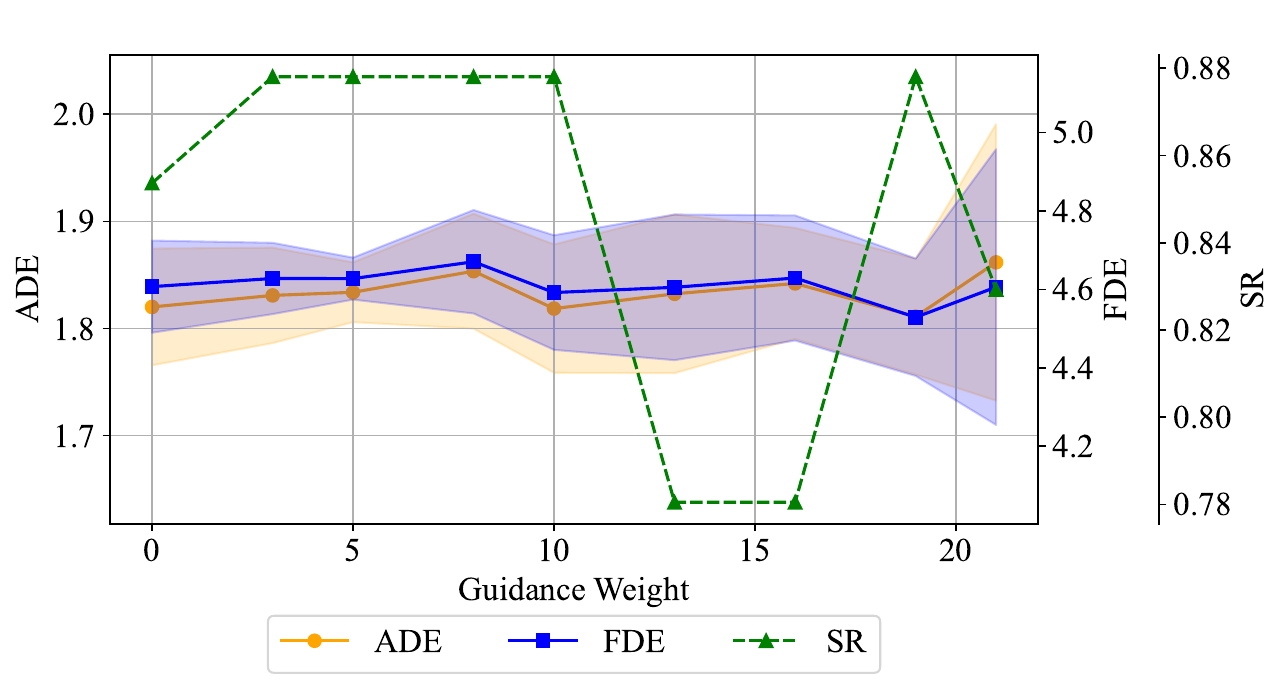}
        \label{fig:GL_w/o}}
    
    
    \subfigure[\textcolor{black}{Diffuser w/ Goals}]{
        \includegraphics[width=0.93\linewidth]{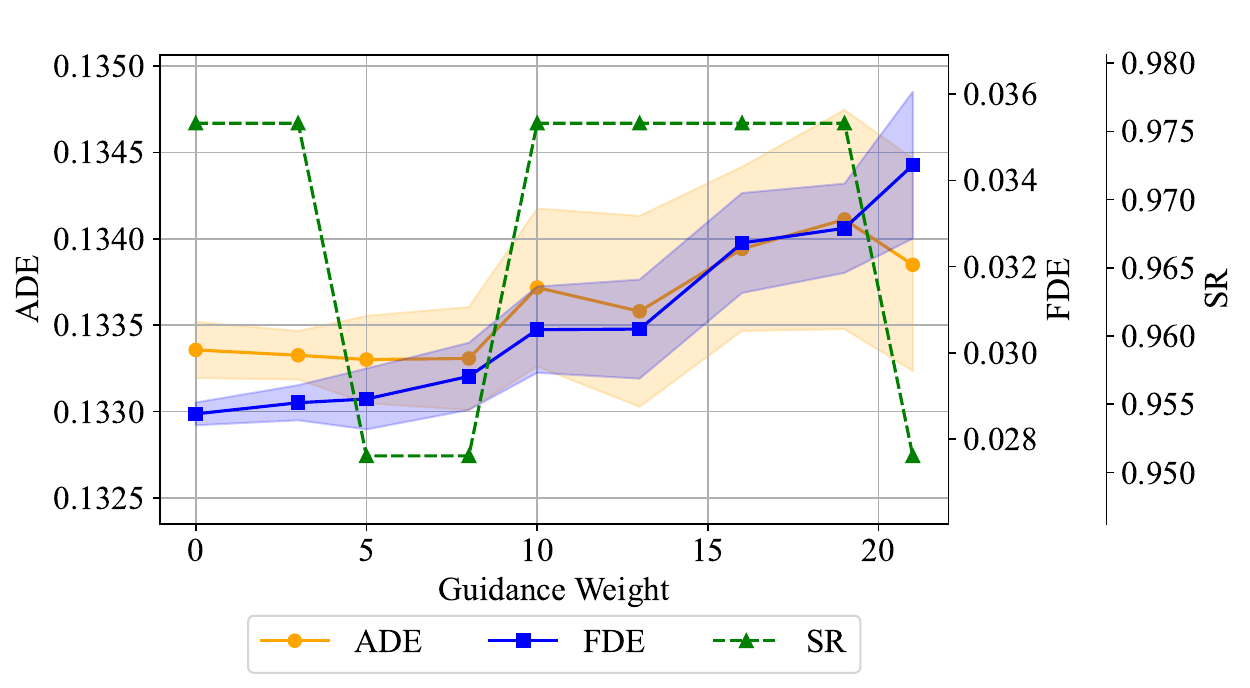}
        \label{fig:GL_w}}
    \caption{\textcolor{black}{Impact of Guidance Weight on Model Performance in GL}}
    \label{fig:GL}
\end{figure}

\begin{figure}[t]
    \centering
    \subfigure[\textcolor{black}{Diffuser w/o Goals}]{
        \includegraphics[width=0.9\linewidth]{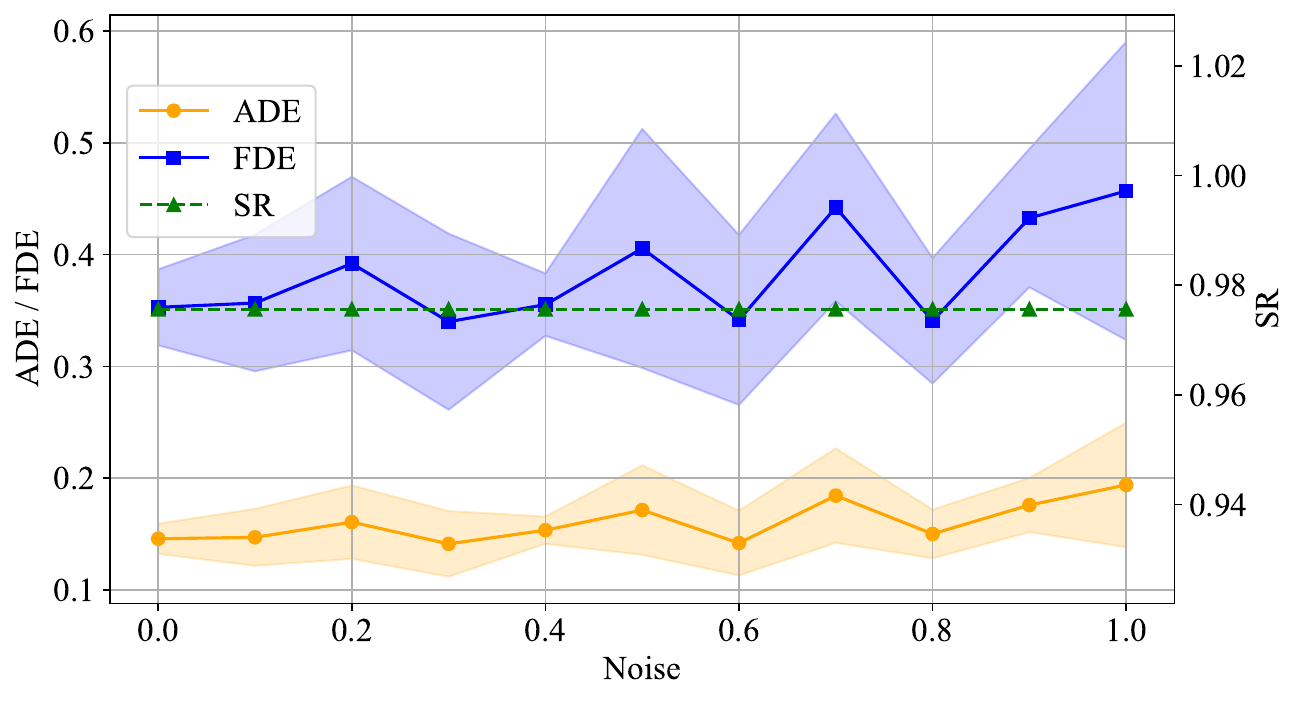}
        \label{noise_MA_w/o}}
    
    
    \subfigure[\textcolor{black}{Diffuser w/ Goals}]{
        \includegraphics[width=0.93\linewidth]{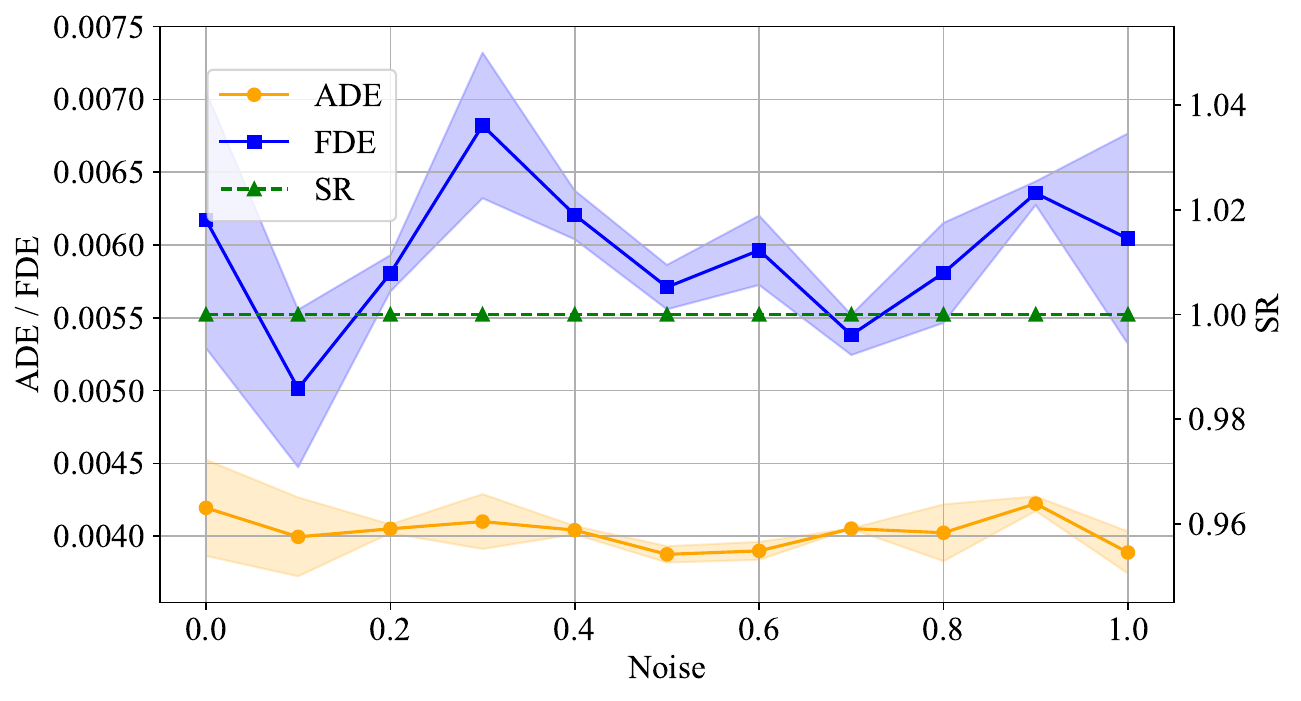}
        \label{noise_MA_w}}
    \caption{\textcolor{black}{Impact of Noise Variance on Model Performance in MA}}
    \label{noise_MA}
\end{figure}

\begin{figure}[htbp]
    \centering
    \subfigure[\textcolor{black}{Diffuser w/o Goals}]{
        \includegraphics[width=0.9\linewidth]{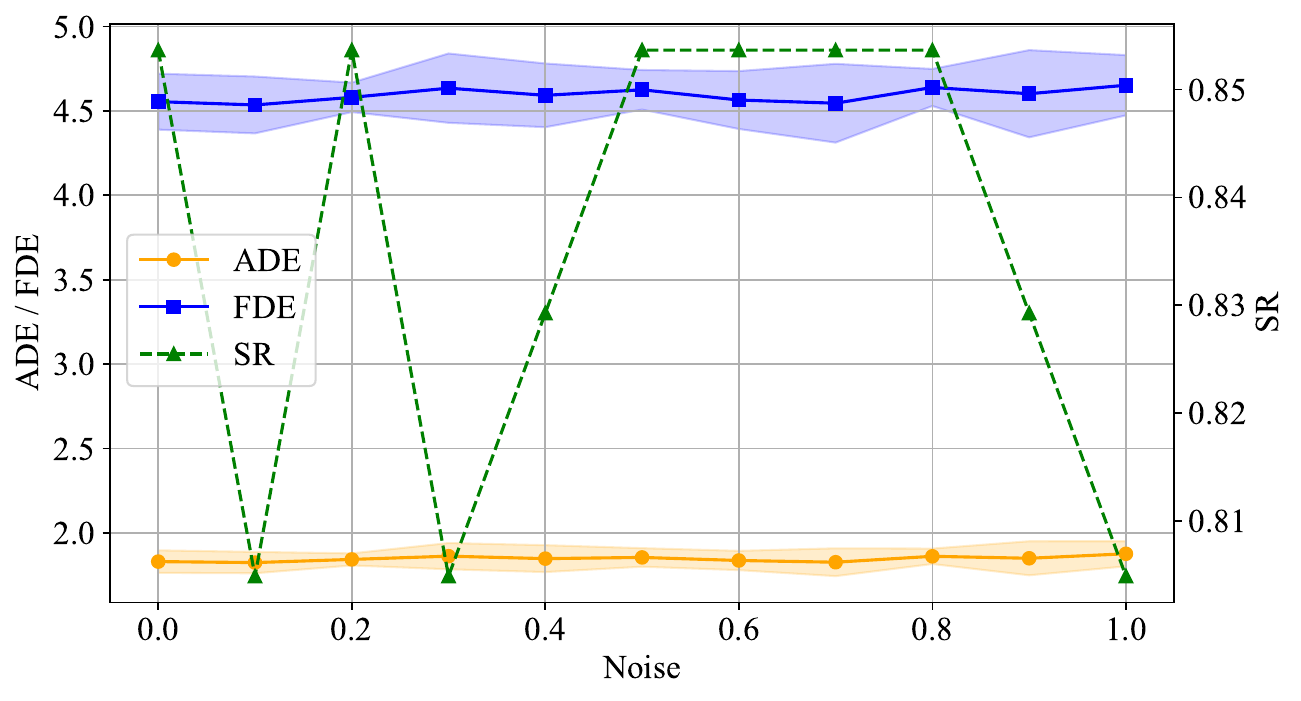}
        \label{noise_GL_w/o}}
    
    
    \subfigure[\textcolor{black}{Diffuser w/ Goals}]{
        \includegraphics[width=0.93\linewidth]{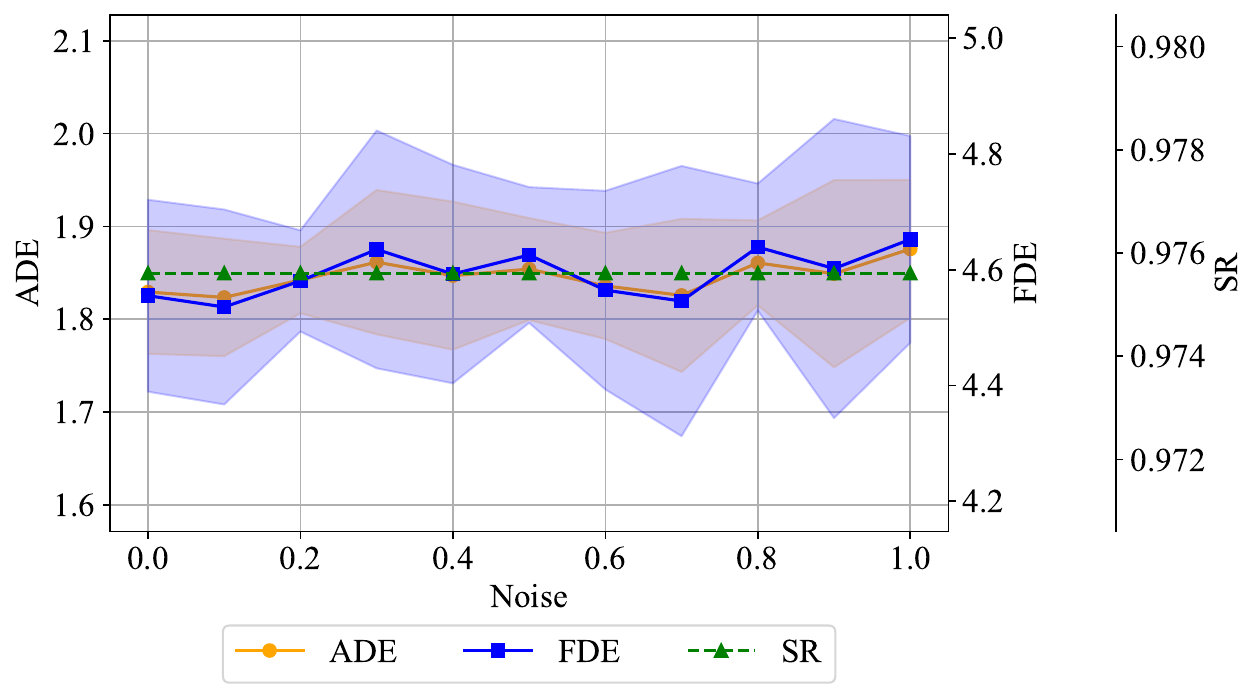}
        \label{noise_GL_w}}
    \caption{\textcolor{black}{Impact of Noise Variance on Model Performance in GL}}
    \label{noise_GL}
\end{figure}
\textcolor{black}{Guidance and goals demonstrate consistent effects across both MA and GL scenarios. In these cases, appropriate levels of guidance significantly improve the model’s trajectory prediction accuracy by reducing ADE and FDE values and enhancing success rates concurrently. For instance, in the MA scenario (Fig. \ref{fig:MA_w/o}), as the guidance weight increases from lower to moderate levels (around a weight of 8), the model consistently improves in performance. In the GL scenario (Fig. \ref{fig:GL_w/o}), with a guidance weight of 18, the model achieves optimal performance. Nevertheless, inappropriate guidance leads to diminishing returns, as indicated by increasing ADE and FDE values along with a decline in success rates. This indicates that although proper guidance aids in refining the task objective, excessively high guidance settings may introduce instability, thereby impairing the model's generalization ability.}

\textcolor{black}{In contrast, when explicit goal information is provided, excessive guidance adversely affects performance in both MA (Fig. \ref{fig:MA_w}) and GL (Fig. \ref{fig:GL_w}) scenarios. Under these goal-informed conditions, the model consistently achieves high performance while sustaining low error rates and high success rates without requiring significant external guidance. Rather than further improving accuracy, higher guidance levels result in slight degradation of ADE, FDE, and success rate metrics. This effect arises because the explicit goal already encodes sufficient information about task objectives, making additional guidance redundant and potentially counterproductive. Therefore, explicitly defined goals effectively ensure stable and robust model behavior, even in untrained scenarios, reducing dependence on external parameter adjustments.}

\textcolor{black}{Without explicit goals (Fig. \ref{noise_MA_w/o} and Fig. \ref{noise_GL_w/o}), the model exhibits high sensitivity to increasing noise levels. In both the MA and GL scenarios, ADE and FDE values fluctuate depending on noise levels. However, overall, setting the noise level to 0 removes sampling entropy, leading to a decline in performance. On the other hand, setting the noise level to 1 increases variance in generated trajectories. When explicit goal information is introduced in both MA and GL scenarios, Fig.\ref{noise_MA_w} and Fig.\ref{noise_GL_w} illustrate significantly more robust model performance. ADE and FDE remain stable, and the SR remain consistently high with minimal fluctuations. This robustness reinforces the idea that explicit goals inherently enhance model stability.}

\section{CONCLUSION}
\label{Conclusion}
\textcolor{black}{This paper proposes a unified multi-task learning and safety-critical planning framework that utilizes a diffusion model to recover policies from expert demonstrations across diverse driving tasks, including turning left, going straight and turning right movements at signal-free intersections. To ensure safe and adaptive trajectory planning, the proposed approach incorporates DHOCBFs as hard constraints, effectively balancing safety and efficiency in dynamic traffic environments. The primary findings and conclusions are summarized as follows:}
\begin{enumerate}
    \item \textcolor{black}{Unlike traditional HOCBF, the DHOCBF exhibits greater flexibility and adaptability in handling dynamic environments. This approach effectively mitigates excessive conservatism, allowing for safer yet more efficient trajectory planning in response to varying obstacle speeds, sizes, uncertainties, and positions.}
    \item \textcolor{black}{In the absence of a goal, task-guided planning enhances efficiency by leveraging appropriate guidance weights to optimize model performance. However, excessive guidance can lead to diminishing returns and overfitting, which highlights the importance of balanced task guidance for optimal trajectory generation.}
    \item \textcolor{black}{Explicit goal conditioning improves trajectory stability, realism, and generalization. By incorporating goal constraints, the DSC-Diffuser encourages vehicles reach their designated lanes, thereby minimizing displacement errors (ADE and FDE) and improving intersection navigation.}
    \item \textcolor{black}{Task guidance has minimal impact on the DSC-Diffuser in the presence of existing goals, indicating that once a goal is defined, the system can autonomously infer tasks, thereby reducing reliance on external guidance and improving planning efficiency.}
    \item \textcolor{black}{Incorporating DHOCBF guarantees driving safety (achieving SR=1) without compromising performance under both trained and untrained scenarios. This ensures trajectory feasibility, stability, and robust safety enforcement, particularly in untrained environments, thereby reinforcing the reliability of the DSC-Diffuser.}
\end{enumerate}

The experiment conducted in this study demonstrates the stability, realism, and generalization of our proposed DSC-Diffuser planning method while maintaining driving safety. Our proposed method offers a promising approach for safe and adaptable autonomous driving. \textcolor{black}{For example, it can be directly applied to the urban deployment of AVs at uncontrolled intersections, where unpredictable traffic behaviors require robust safety measures. Moreover, integration with existing traffic management systems can optimize traffic flow, reduce congestion, and enhance overall road safety in complex urban environments.} There exist some limitations. Our study assumes that other vehicles do not respond to the ego vehicle's actions, which is a significant limitation in autonomous driving planning \cite{rhinehart2021contingencies}. \textcolor{black}{It is also assumed that the sensory system can perfectly obtain the states of other vehicles, which could be addressed by integrating methods that consider prediction uncertainty, sensor noise, and communication delays \cite{gao2020feature,ding2023end}. Furthermore, this study did not account for variations in vehicle perception and control due to the lack of fully developed models capturing the diverse capabilities of real-world vehicles, which will be considered in future research. For future work, we aim to explore the dynamic selection of DHOCBF parameters to further improve the adaptability and then expend our methods to connected autonomous vehicles \cite{mirheli2019consensus,wang2023coordination,gong2024collision}.}

\small
\bibliography{ref}

\begin{IEEEbiography}[{\includegraphics[width=1in,height=1.35in,clip,keepaspectratio]{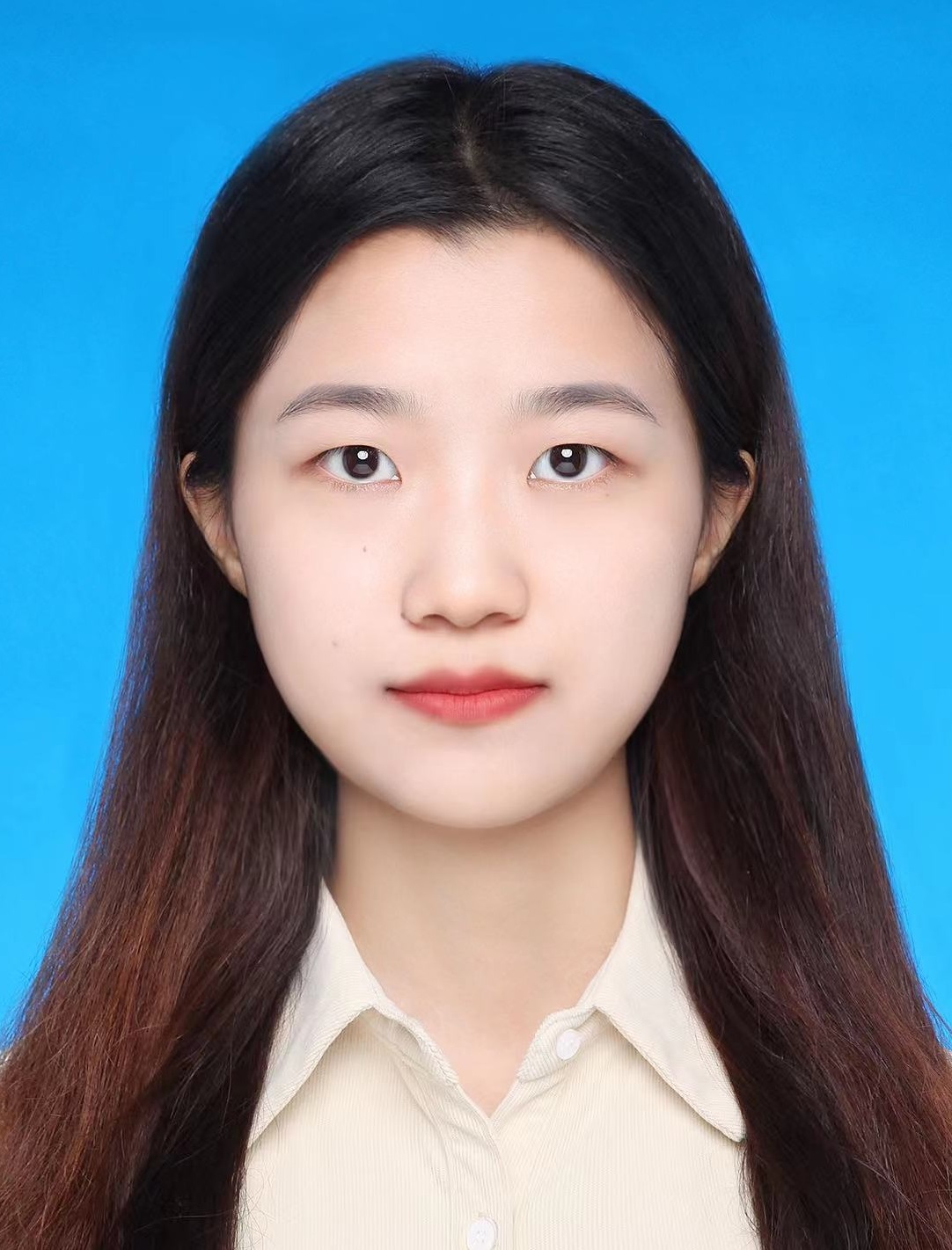}}]{Di Chen} received the B.Sc. and M.S. degree in Transportation Engineering from School of Transportation engineering, Tongji University, Shanghai, China. She is a research assistant in the Department of Electrical and Electronic Engineering of The Hong Kong Polytechnic University. Her research interests include traffic simulation, machine learning methods for intelligent transportation.
\end{IEEEbiography}
\begin{IEEEbiography}[{\includegraphics[width=1in,height=1.35in,clip,keepaspectratio]{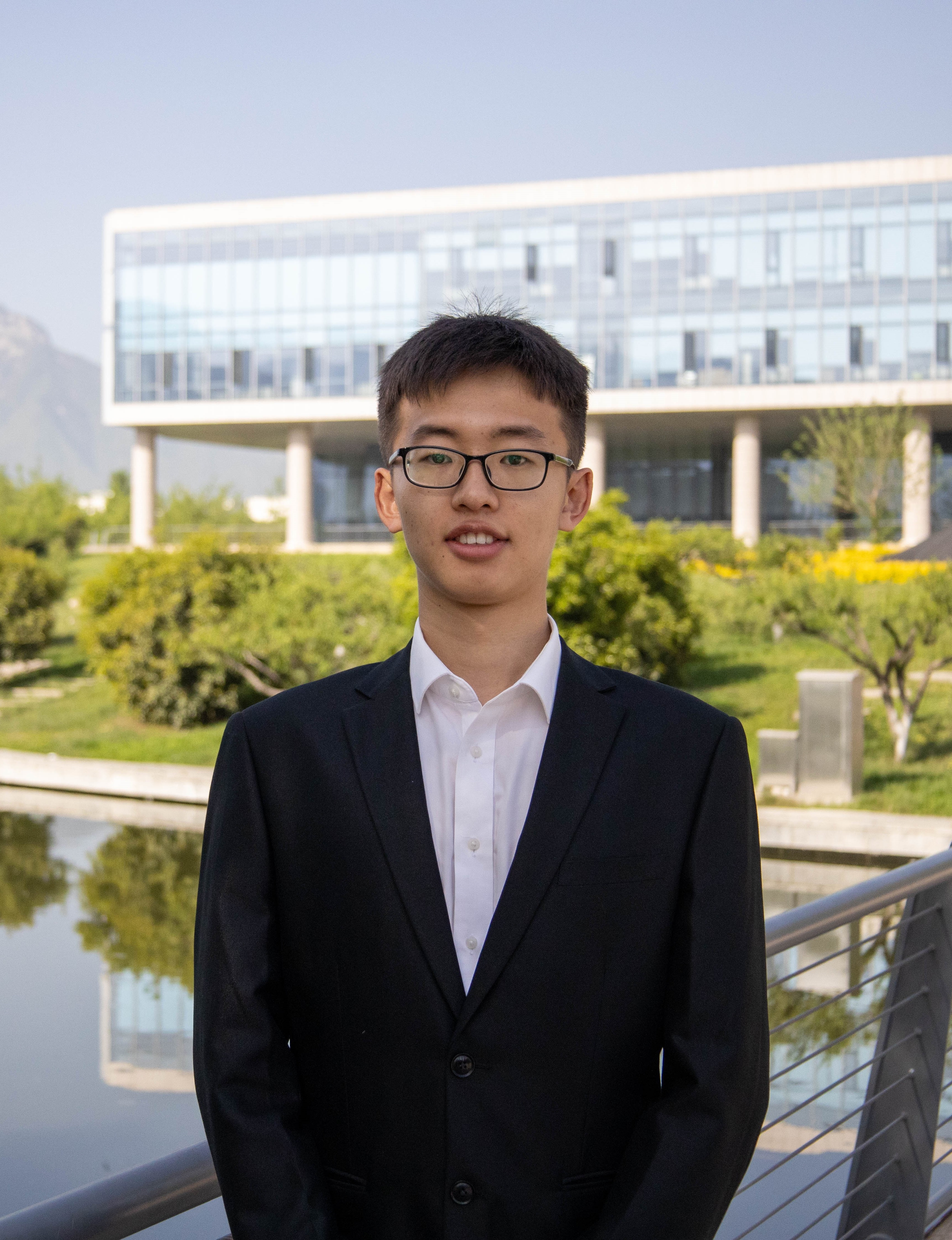}}]{Ruiguo Zhong} received the B.Sc. and M.S. degree in Control Science and Engineering from School of Electronics and Information, Northwestern Polytechnical University, Shaanxi, China. Currently, he is a Ph.D. student in the Hong Kong University of Science and Technology (Guangzhou), China. His research interests include generative methods for intelligent transportation.
\end{IEEEbiography}
\begin{IEEEbiography}[{\includegraphics[width=1in,height=1.35in,clip,keepaspectratio]{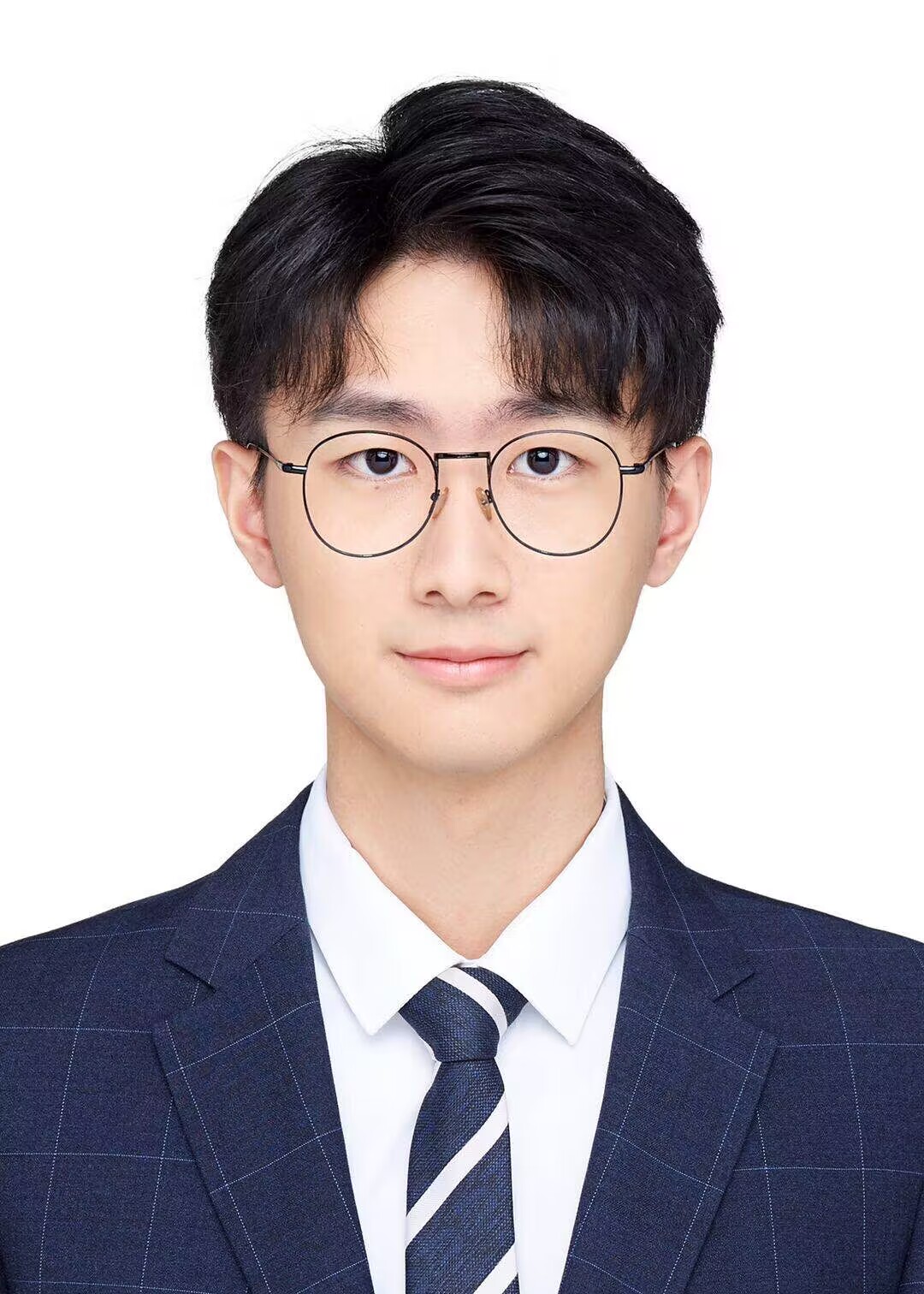}}]{Kehua Chen} received a B.S. degree in Civil Engineering from Chongqing University and a dual M.S. degree in Environmental Sciences from the University of Chinese Academy of Sciences and the University of Copenhagen. He earned his Ph.D. in Intelligent Transportation from the Hong Kong University of Science and Technology in 2024. Currently, he is a postdoctoral scholar at the Smart Transportation Applications and Research (STAR) Lab at the University of Washington. His research interests encompass urban and sustainable computing, as well as autonomous driving.
\end{IEEEbiography} 
\begin{IEEEbiography}
[{\includegraphics[width=1in,height=1.35in,clip,keepaspectratio]{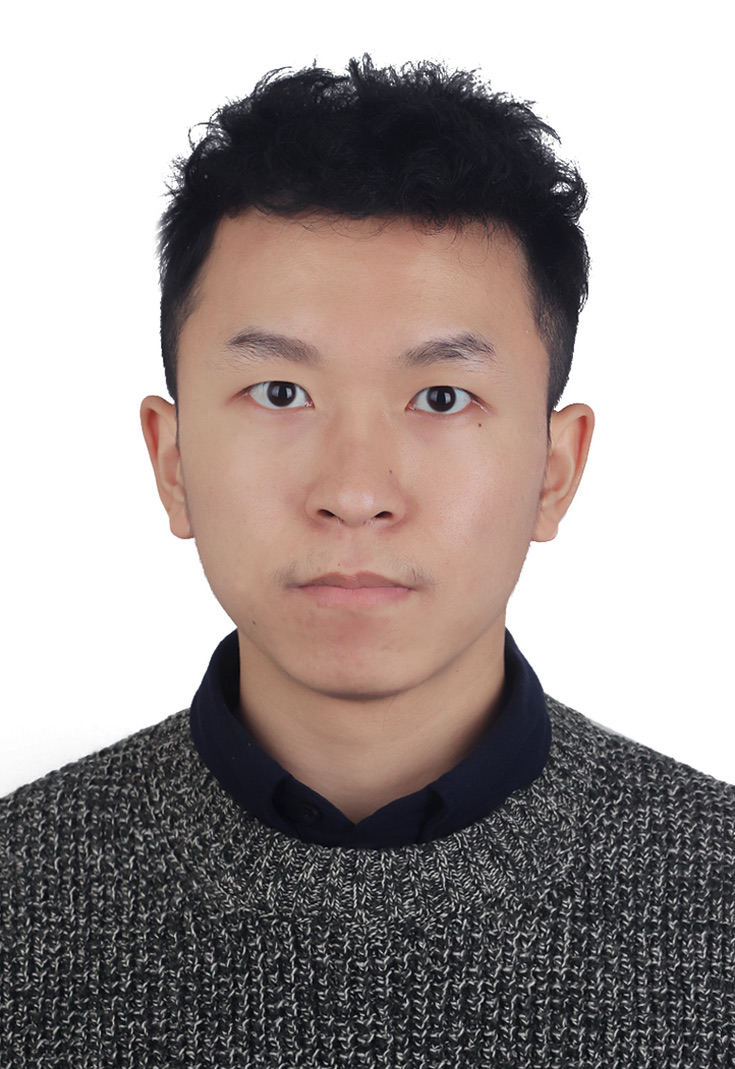}}]{Zhiwei Shang } received the B.S. degree in Hydrology and Water Resources Engineering from the Sichuan University, China, in 2020, and the M.S.degree in Computer Technology from University of Chinese Academy of Sciences, China, in 2023. Currently, he is a research assistant at the Hong Kong University of Science and Technology (Guangzhou), China. His research interests mainly include reinforcement learning and intelligent control.
\end{IEEEbiography}

\begin{IEEEbiography}[{\includegraphics[width=1in,height=1.35in,clip,keepaspectratio]{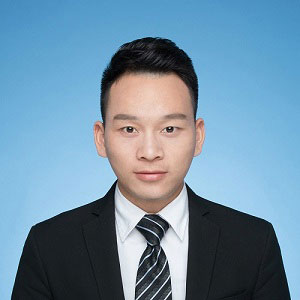}}]{Meixin Zhu} is a tenure-track Assistant Professor in the Thrust of Intelligent Transportation (INTR) under the Systems Hub at the Hong Kong University of Science and Technology (Guangzhou). He is also an affiliated Assistant Professor in the Civil and Environmental Engineering Department at the Hong Kong University of Science and Technology. He obtained a Ph.D. degree in intelligent transportation at the University of Washington (UW) in 2022. He received his B.S. and M.S. degrees in traffic engineering in 2015 and 2018, respectively, from Tongji University. His research interests include Autonomous Driving Decision Making and Planning, Driving Behavior Modeling, Traffic-Flow Modeling and Simulation, Traffic Signal Control, and (Multi-Agent) Reinforcement Learning. He is a recipient of the TRB Best Dissertation Award (AED50) in 2023.
\end{IEEEbiography}

\begin{IEEEbiography}[{\includegraphics[width=1in,height=1.35in,clip,keepaspectratio]{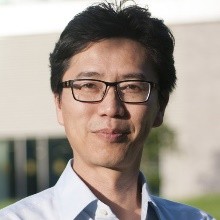}}]{Edward Chung} is a Professor of Intelligent Transport Systems (ITS) at the Department of Electrical Electronic Engineering of The Hong Kong Polytechnic University. Edward received a Bachelor of Civil Engineering with Honours and PhD from Monash University. With an extensive background as both an engineer and an accomplished academic researcher, Edward has garnered significant experience working on national and international projects. Notably, he held positions such as Senior Research Scientist at the Australian Road Research Board, Manager of Infrastructure Analysis and Modelling at the Victorian Department of Infrastructure, Australia, Visiting Professor at the Centre for Collaborative Research, University of Tokyo, and Head of the ITS Group at LAVOC, EPFL, Switzerland. Prior to his current role at PolyU, Edward served as a professor at the Queensland University of Technology (QUT) and held the position of Director of the Smart Transport Research Centre at QUT.
\end{IEEEbiography}
\end{CJK}
\end{document}